\documentclass[10pt,twocolumn,letterpaper]{article}

\usepackage[pagenumbers]{wacv} 

\usepackage[table,xcdraw]{xcolor}
\usepackage{graphicx}
\usepackage{amsmath}
\usepackage{amssymb}
\usepackage{booktabs}
\usepackage{comment}
\usepackage{multirow}
\usepackage{xr}
\usepackage{algorithm,algorithmic}
\usepackage[toc,page]{appendix}

\usepackage[pagebackref,breaklinks,colorlinks]{hyperref}

\usepackage[capitalize]{cleveref}
\crefname{section}{Sec.}{Secs.}
\Crefname{section}{Section}{Sections}
\Crefname{table}{Table}{Tables}
\crefname{table}{Tab.}{Tabs.}

\def\wacvPaperID{1072} 
\def\confName{WACV}
\def\confYear{2024}

\begin{document}

\title{Domain Generalization by Rejecting Extreme Augmentations}

\author{Masih Aminbeidokhti\\
{\tt\small masih.aminbeidokhti.1@ens.etsmtl.ca}
\and
Fidel A. Guerrero Pe\~{n}a\\
{\tt\small fidel-alejandro.guerrero-pena@etsmtl.ca}
\and
Heitor Rapela Medeiros\\
{\tt\small heitor.rapela-medeiros.1@ens.etsmtl.ca}
\and
Thomas Dubail\\
{\tt\small thomas.dubail.1@ens.etsmtl.ca}
\and
Eric Granger\\
{\tt\small eric.granger@etsmtl.ca}
\and
Marco Pedersoli\\
{\tt\small marco.pedersoli@etsmtl.ca}
\and
LIVIA, Dept. of Systems Engineering\\
ETS Montreal, Canada
}

\maketitle


\begin{abstract}
Data augmentation is one of the most effective techniques for regularizing deep learning models and improving their recognition performance in a variety of tasks and domains. However, this holds for standard in-domain settings, in which the training and test data follow the same distribution. For the out-of-domain case, where the test data follow a different and unknown distribution, the best recipe for data augmentation is unclear. In this paper, we show that for out-of-domain and domain generalization settings, data augmentation can provide a conspicuous and robust improvement in performance. To do that, we propose a simple training procedure:
(i) use uniform sampling on standard data augmentation transformations; 
(ii) increase the strength transformations to account for the higher data variance expected when working out-of-domain; and (iii) devise a new reward function to reject extreme transformations that can harm the training. 
With this procedure, our data augmentation scheme achieves a level of accuracy that is comparable to or better than state-of-the-art methods on benchmark domain generalization datasets. Code: \url{https://github.com/Masseeh/DCAug}

\end{abstract}

\section{Introduction}

\begin{figure}[t]
   \centering
    \includegraphics[width=\linewidth]
    {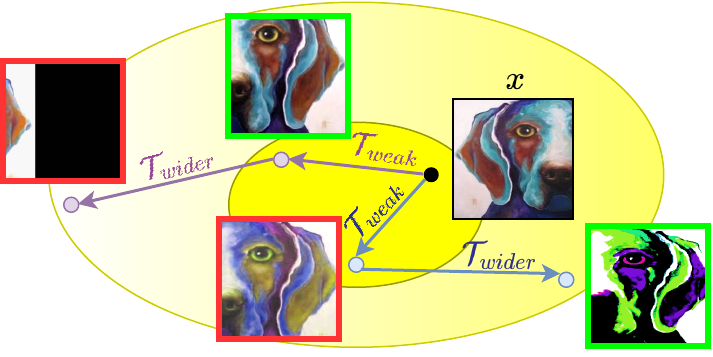}
 
    \caption{A conceptual illustration of our method. 
    The inner circle and outer circle represent the space of weak (safe) and wider (possibly harmful) augmentations, respectively. Our method is able to automatically select for each combination of data samples and augmentation a wider transformation (when safe) or reject it when unsafe. This is achieved with the help of a reward function (represented as the yellow color gradient) that compares the diversity and the consistency of an augmented sample (see Section \ref{sec:4} for more details).
    In the illustration, given an image $x$, we present two possible paths of augmentation. For the blue path, the wide augmentation has a high diversity and high consistency, and therefore it is selected (green box). For the purple path, although the wide augmentation has high diversity, it also has low consistency, therefore the transformation is rejected (red box), and the weak transformation is used instead as augmentation.} 
    \label{fig:onecol}
 \end{figure}

 
The main assumption of commonly used deep learning methods is that all examples used for training and testing models are independently and identically sampled from the same distribution \cite{vapnik1991principles}. In practice, such an assumption does not always hold and this can limit the applicability of the learned models in real-world scenarios \cite{koh2021wilds}. 

In order to tackle this problem, domain generalization (DG) \cite{blanchard2011generalizing} aims to predict well data distributions different from those seen during training. In particular, we assume access to multiple datasets during training, each of them containing examples about the same task but collected under a different domain or environment. One effective approach for DG is to increase the diversity of the training data \cite{yang2022survey}. Data augmentation, which is a widely used approach for generating additional training data \cite{krizhevsky2017imagenet, szegedy2015going, park2019specaugment}, is especially beneficial since it can help to approximate the true distribution of the dataset. However, choosing augmentations often depends on the underlying dataset. While learning the right data augmentation for in-domain setting (same distribution between training and test) has been explored in research \cite{cubuk2019autoaugment, cubuk2020randaugment, muller2021trivialaugment, suzuki2022teachaugment}, there is currently little research on good augmentations for domain generalization and how to leverage domain information to make the augmentations more effective.  

In this work, we investigate those questions. First, we show that data augmentation is also useful for domain generalization, but to cover the different training domains and hopefully the target domain the proposed transformations need to be stronger than for in-domain tasks. However, too strong transformations could be harmful to the learning process (see Figure \ref{fig:onecol}). To fully exploit stronger transformations, without harming the learning, we select diverse and challenging samples that provide helpful training information without losing the sample's original semantics. We introduce a reward function consisting of diversity and semantic consistency components and use it to select for each sample the best augmentation between a weak but safe and a strong but diverse augmentation. Thus, the proposed algorithm should be able to select which augmentation is better for each sample for training a model that can generalize to unknown data distributions. 

\noindent \textbf{The main contributions of this paper are as follows:}

\noindent \textbf{(1)} We show that while commonly used augmentation-based techniques for in-domain settings are quite powerful for DG, we can increase the performance further by expanding the range of transformations. Consequently, we achieve superior results compared to the majority of approaches relying on domain-invariant representation.
    
\noindent \textbf{(3)} With the new expanded range is easier to produce harmful transformations, therefore we introduce a data augmentation schema that selects the optimal augmentation strategy between a weak yet safe and a diverse yet strong augmentation technique.
    
\noindent \textbf{(3)} Experimental on common benchmark datasets show the benefits of our proposed method achieving an accuracy that is better than state-of-the-art methods for DG.


\section{Related Works}
\label{sec:2}

\textbf{Data Augmentation:} There has been extensive research on data augmentation for computer vision tasks. Horizontal flips and random cropping or translations of images are commonly used for natural image datasets such as CIFAR-10 \cite{krizhevsky2009learning} and ImageNet \cite{russakovsky2015imagenet}, while elastic distortions and scalings are more common on MNIST dataset \cite{yang2022image}. While data augmentation usually improves model generalization, if too strong, it might sometimes hurt performance or induce unexpected biases. Thus, one needs to manually find effective augmentation policies based on domain knowledge and model validation. To alleviate this issue, researchers propose various methods to automatically search efficient augmentation strategies for model in-domain generalization \cite{cubuk2019autoaugment,lim2019fast,hataya2020faster, zhou2020deep, ho2019population, zhang2019adversarial}. AutoAugment (AA) \cite{cubuk2019autoaugment} is the pioneering work on automating the search for the ideal augmentation policy which uses Reinforcement Learning (RL) to search for an optimal augmentation policy for a given task. Unfortunately, this search process requires extensive computing power, to the order of several thousands of GPU hours. Many subsequent works adopt AutoAugment search space for their own policy search \cite{yang2022survey}. In particular, \cite{li2020dada, mounsaveng2021learning} propose methods to shorten the duration of the policy search for data augmentation while maintaining similar performance. Alternatively, other works resort to different guided search techniques to accelerate the search. Lim \etal~\cite{lim2019fast} uses a Bayesian optimization approach, \cite{hataya2020faster} uses an online search during the training of the final model, and \cite{ho2019population} employs an evolutionary algorithm to search for the optimal augmentation policy also in an online fashion. Adversarial AutoAugment (Adv. AA) \cite{zhang2019adversarial} is another slightly cheaper method that uses multiple workers and learns the augmentation policy that leads to hard
samples measured by target loss during training. However, all of these sophisticated approaches are comparable to RandAugment (RA), which uses the augmentation search space introduced in \cite{cubuk2020randaugment}, but with a uniform sampling policy in which only the global magnitude of the transformations and the number of applied transformations are learned on a validation set. TrivialAugment (TA) \cite{muller2021trivialaugment} and UniformAugment \cite{lingchen2020uniformaugment} further push the RA method to the extreme and propose to use a truly search-free approach for data augmentations selection, yet achieving test set accuracies that are on-par or better than the more complex techniques previously discussed. However, all the mentioned methods use the search space of AutoAugment, which is already designed to not excessively distort the input image. To control the space of data augmentation, Gong \etal~\cite{gong2021keepaugment} regularize augmentation models based on prior knowledge while Wei \etal~\cite{wei2020circumventing} use knowledge distillation to mitigate the noise introduced by aggressive AA data augmentation policies. Suzuki~\cite{suzuki2022teachaugment} proposes an online data augmentation optimization method called TeachAugment that introduces a teacher model into the adversarial data augmentation and makes it more informative without the need for careful parameter tuning. However, all the mentioned methods are designed for standard in-domain settings and do not consider the generalization problem for unknown domains as in domain generalization problems. 

\textbf{Domain Generalization:} Learning domain-invariant features from source domains is one of the most popular methods in domain generalization. These methods aim at learning high-level features that make domains statistically indistinguishable (domain-invariant). Ganin \etal~\cite{ganin2016domain} propose Domain Adversarial Neural Networks (DANN), which uses GAN, to enforce that the features cannot be predictive of the domain. Albuquerque \etal~\cite{albuquerque2019generalizing} build on top of DANN by considering one-versus-all adversaries that try to predict to which training domain each of the examples belongs. Later work, consider a number of ways to enforce invariance, such as minimizing the maximum mean discrepancy (MMD) \cite{li2018domain}, enforcing class-conditional distributions across domains \cite{li2018deep}, and matching the feature covariance (second order statistics) across training domains at some level of representation \cite{sun2016deep}. Although popular, enforcing invariance is challenging and often too restrictive. As a result, Arjovsky \etal~\cite{arjovsky2019invariant} propose to enforce the optimal classifier for different domains. GroupDRO \cite{sagawa2019distributionally} proposes to minimize the worst-case training loss by putting more mass on samples from the more challenging domains at train time. Bui \etal~\cite{bui2021exploiting} use meta-learning and adversarial training in tandem to disentangle features in the latent space while jointly learning both domain-invariant and domain-specific features in a unified framework. However, Zhao \etal~\cite{zhao2019learning} show that learning an invariant representation, in addition to possibly ignoring signals that can be important for new domains, is not enough to guarantee target generalization. Furthermore as evidenced by the strong performance of ERM \cite{gulrajani2020search}, these methods are either too strong to optimize reliably or too weak to achieve their goals \cite{zhang2022rich}.

\textbf{Data Augmentation for Domain Generalization:} Another effective strategy to address domain generalization \cite{wiles2021fine,gulrajani2020search} is by using data augmentation. These methods focus on manipulating the inputs to assist in learning general representations. Zhou \etal~\cite{zhou2020deep} use domain information for creating an additive noise to increase the diversity of training data distribution while preserving the semantic information of data. Yan \etal~\cite{yan2020improve} use mixup to blend examples from the different training distributions. In RSC \cite{huang2020self}, the authors iteratively discard the dominant features from the training data, aiming to improve generalization. This approach is inspired by the style transfer literature, where the feature statistics encode domain-related information. Similarly, MixStyle \cite{zhou2021domain} synthesizes novel domains by mixing the feature statistics of two instances. SagNets \cite{nam2021reducing} propose to disentangle style encodings from class categories to prevent style-biased predictions and focus more on the contents. The performance of these methods depends on whether the augmentation can help the model to learn invariance in the data. 

In this work, we build upon observations from \cite{wiles2021fine, zhang2019unseen, gulrajani2020search}, which show that data augmentation plays a vital role in improving out-of-distribution generalization. Our approach employs uniform sampling, similar to TrivialAugment \cite{muller2021trivialaugment}, and a rejection reward inspired by TeachAugment \cite{suzuki2022teachaugment}. This combination leads to the proposal of an effective data augmentation strategy for domain generalization. 

\section{Revisiting Random Data Augmentation for Domain Generalization}
\label{sec:3}

\subsection{Problem Definition}

We study the problem of Multi-Source Domain Generalization for classification. During training, we assume access to $N$ datasets containing examples about the same task but collected under a different domain or environment, $\mathcal{D} = \{1,2,..,N\}$. Let $\mathcal{S}$ be a training dataset containing samples from all training domains, $\mathcal{S} = \{(x_1, y_1, d_1), (x_2, y_2, d_2) \dots, (x_M, y_M, d_M) \}$, with $M = |\mathcal{S}|$. Here, $x_i\in\mathcal{X}$ refers to an image, $y_i\in\mathcal{Y}$ is the class label, and $d_i\in\mathcal{D}$ is the domain label.  
Then, the goal of the domain generalization task is to learn a mapping $f_\theta\colon\mathcal{X}\to\mathcal{Y}$ parametrized by $\theta$ that generalizes well to an unseen domain, $\hat{d}\notin\mathcal{D}$. In addition, we also consider a domain classifier $h_{\phi}\colon\mathcal{X}\to\mathcal{D}$ parametrized by $\phi$ that learns to recognize the domain of a given sample from $\mathcal{S}$. As a baseline optimization problem, we consider the simple empirical risk minimization (ERM), which minimizes the average loss over all samples, $\theta^* = \arg\min_\theta \frac{1}{M}\sum_{(x,y)\in\mathcal{S}} \mathcal{L}(f_\theta(x), y)$, where $\mathcal{L}(\cdot)$ is the cross-entropy loss function.

\subsection{Data Augmentation Search Space}


A well-known approach to achieving domain generalization is transforming the training samples during the learning process to gain robustness against unseen domains \cite{cubuk2019autoaugment, muller2021trivialaugment}. These transformations come from a predefined set of possible data augmentations that operate within a given range of magnitudes. We consider as standard transformations $\mathcal{T}_{weak}\colon\mathcal{X}\to\mathcal{X}_{weak}$ the random flip, crop, and slight color-jitter augmentations that are safe, i.e., do not destroy image semantics. Such weak transformations are used in every training step. On top of standard transformation, we may also apply more transformations selected from either data augmentation search space Default from RandAugment \cite{cubuk2020randaugment} or Wide from TrivialAugment (TA) \cite{muller2021trivialaugment}. Here we use geometric transformations (ShearX/Y, TranslateX/Y, Rotate) as well as color-based transformations (Posterize, Solarize, Contrast, Color, Brightness, Sharpness, AutoContrast, Equalize, and Grey). However, unlike TA, we expand the magnitude ranges and construct $\mathcal{T}_{wider}\colon\mathcal{X}_{weak}\to\mathcal{X}_{wider}$ to include more aggressive data augmentation (see \cref{supp-tab:space} from supplementary materials). For sampling transformations, we follow the TA procedure, which involves randomly sampling an operation and magnitude from the search space for each image. 

\begin{table*}[h]
    \centering
    \begin{tabular}{@{}lccccccc@{}}

        \toprule
        \multirow{2}{*}[-1em]{\textbf{Method}} & \multirow{2}{*}[-1em]{\textbf{Search Space}} & {} & {} & {} &
        \\
        
        {} & {} & {} & \multicolumn{3}{c}{\multirow{2}{*}[1em]{$\qquad$$\qquad$\textbf{Dataset}}} \\

        \cmidrule(lr){3-8}
        \addlinespace[5pt]

        {} & {} & PACS & VLCS & OfficeHome & TerraInc & DomainNet & \textbf{Avg.}\\
        
        \midrule

        ERM as in~\cite{vapnik1991principles} &
        {Weak} &
        84.2$\scriptstyle\pm 0.1$  &
        77.3$\scriptstyle\pm 0.1$  &
        67.6$\scriptstyle\pm 0.2$  &
        47.8$\scriptstyle\pm 0.6$  &
        44.0$\scriptstyle\pm 0.1$  &
        64.2  \\

        RandAugment \cite{cubuk2020randaugment} &
        {Default} &
        86.1$\scriptstyle\pm 0.8$  &
        78.7$\scriptstyle\pm 0.7$   &
        67.8$\scriptstyle\pm 0.4$   &
        44.7$\scriptstyle\pm 1.4$  &
        44.0$\scriptstyle\pm 0.2$   &
        64.3 \\
        
        TA \cite{muller2021trivialaugment} &
        {Wide} &
        85.5$\scriptstyle\pm 1.1$  &
        78.6$\scriptstyle\pm 0.5$  &
        68.0$\scriptstyle\pm 0.2$  &
        47.8$\scriptstyle\pm 1.6$  &
        43.8$\scriptstyle\pm 0.2$  &
        64.7  \\
        
        AutoAugment \cite{cubuk2019autoaugment} &
        {Default} &
        85.8$\scriptstyle\pm 0.5$  &
        78.7$\scriptstyle\pm 0.8$   &
        68.4$\scriptstyle\pm 0.2$   &
        48.0$\scriptstyle\pm 1.3$  &
        43.7$\scriptstyle\pm 0.2$   &
        64.9 \\
        
        TA (Ours) &
        {Wider} &
        85.6$\scriptstyle\pm 0.8$  &
        78.6$\scriptstyle\pm 0.4$  &
        68.9$\scriptstyle\pm 0.4$  &
        48.3$\scriptstyle\pm 0.8$  &
        43.7$\scriptstyle\pm 0.3$  &
        65.0  \\

        \bottomrule
    \end{tabular}
\caption{Different strategies of data augmentation. We compare different search space ranges traditionally used and wider ones. TA with a wider search space leads to better average out-of-domain accuracy. Our experiments are repeated three times. For details about datasets and training procedures see section \ref{sec:5}.}

\label{tab:motivation}
\end{table*}

\subsection{Motivation for Wider Range}

\begin{figure}[h]
   \centering
    \includegraphics[width=\linewidth]{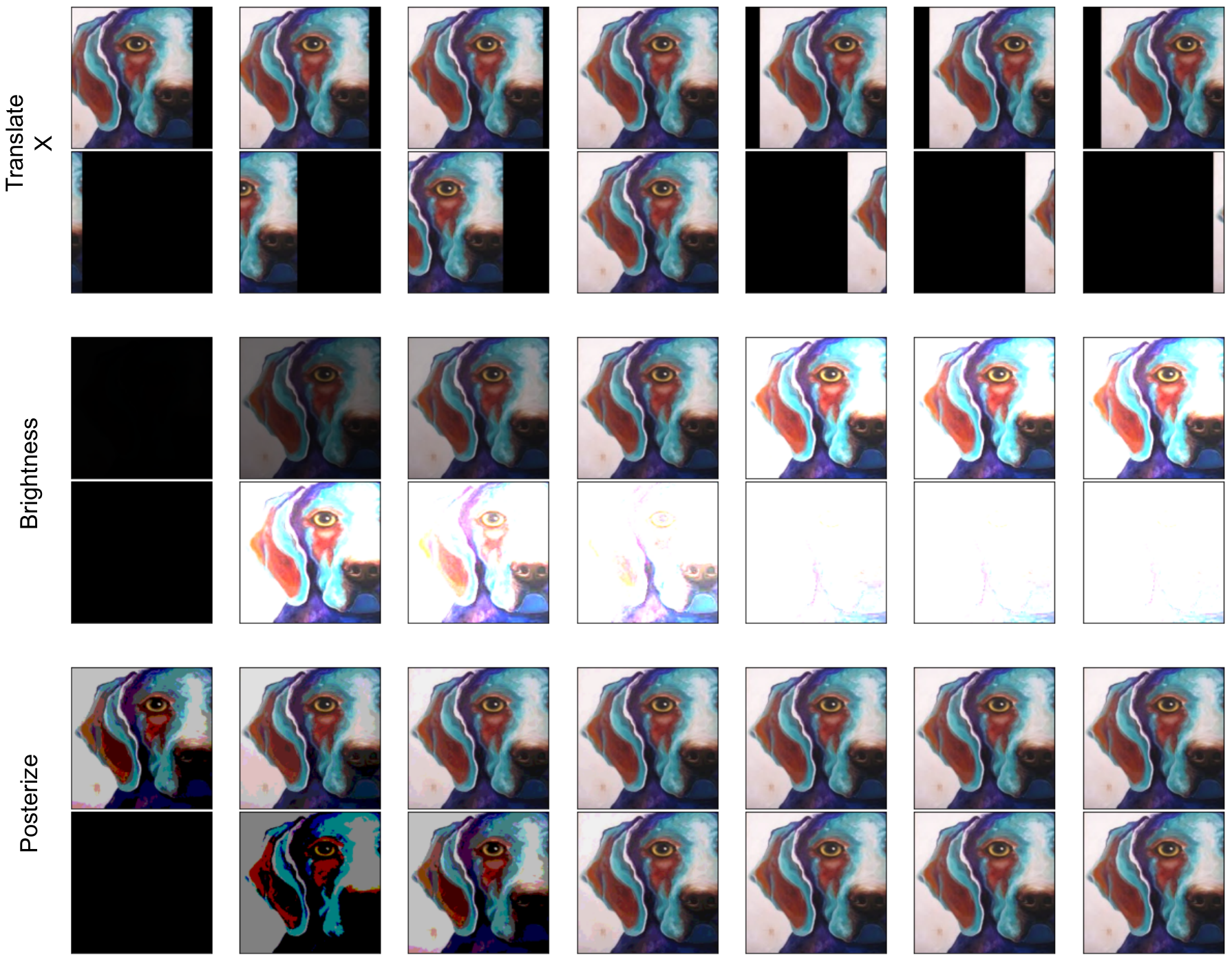}
 
    \caption{Sample transformations from TA with wide and wider search space on PACS dataset. For each transformation, the first row shows the range of transformed samples with wide search space and the second row with wider. We see that the wider space can lead to more variety but also extreme and detrimental transformations that do not keep the semantics of the image. This motivates us to use wider transformations but find a way to reject the extreme ones. 
    }
    \label{fig:augs}
 \end{figure}

Random augmentation over a set of predefined transformations as in TA, despite being very simple, is competitive to the state-of-the-art Data Augmentation in standard in-domain settings. 
In Table \ref{tab:motivation}, we consider the performance of TA, RandAugment, and AutoAugment for domain generalization. As can be seen, such DA methods are already improving over ERM\cite{vapnik1991principles}. However, for domain generalization, we expect that more aggressive transformations can push the representation outside the training domains and help to adapt to new domains.
In fact, as shown in Table \ref{tab:motivation}, the uniform sampling strategy of TA, but with wider transformations further improves over the rest of the methods. 
However, as shown in Figure \ref{fig:augs}, stronger augmentations can easily lead to extreme transformations that do not keep the semantics of the image. Thus, the aim of this work is to further improve this strong baseline by proposing a mechanism to reject those extreme augmentations. For more details about the used datasets and the training procedures, see section \ref{sec:5}.

\section{Rejecting Extreme Transformations} \label{sec:4}
 For each given input, we generate a weakly augmented version using standard transformation (i.e., using only a flip and a crop and slight color-jitter) and a strongly augmented version using $\mathcal{T}_{wider}$ transformation as defined in the previous section. We then define a reward function $R(x,z)$ that given an input $x$ and metadata, either domain label $d$ or class label $y$, provides a measurement of the quality for the transformed sample. Then, maximizing such a reward function allows selecting which augmentation is more suitable for the training:
\begin{equation} \label{eq:1}
\tilde{x} = 
\left\{
	\begin{array}{ll}
		\mathcal{T}_{wider}(x)  & \mbox{if } R(\mathcal{T}_{wider}(x), z) \geq R(\mathcal{T}_{weak}(x), z) \\
            \mathcal{T}_{weak}(x) & \mbox{otherwise } \\
	\end{array}
\right.
\end{equation}
In the following, we define the reward function used.

\subsection{Augmentation Reward}

Intuitively, for domain generalization, a good augmentation creates challenging samples that provide useful training information without losing the sample's original meaning (i.e., the sample's class). We use the teacher-student paradigm to achieve this goal and introduce a unified reward function consisting of diversity and semantic consistency components for selecting the appropriate augmentation. 

Considering $\tilde{x}$ as an augmented sample, the reward function is defined as:
\begin{equation}
   R(\tilde{x}, z) =  (1-\lambda) R_{div}(\tilde{x}, z) - \lambda R_{con}(\tilde{x}, z) 
\end{equation}
where $\lambda$ is the balancing coefficient between diversity and consistency. Here, $z$ refers to either the domain of the sample $d$ or the class label $y$, and it is specified in the following sections for every term of the proposed reward. In the previous equation, the $R_{div}$ term enforces diversity in the data by exploring the augmentations of the input, while $R_{con}$ keeps the semantic meaning of augmented sample $\tilde{x}$. 

\subsection{Diverse Student and Consistent Teacher}
\label{sec:4:2}

\begin{figure*}[h]
   \centering
    \includegraphics[width=\linewidth]{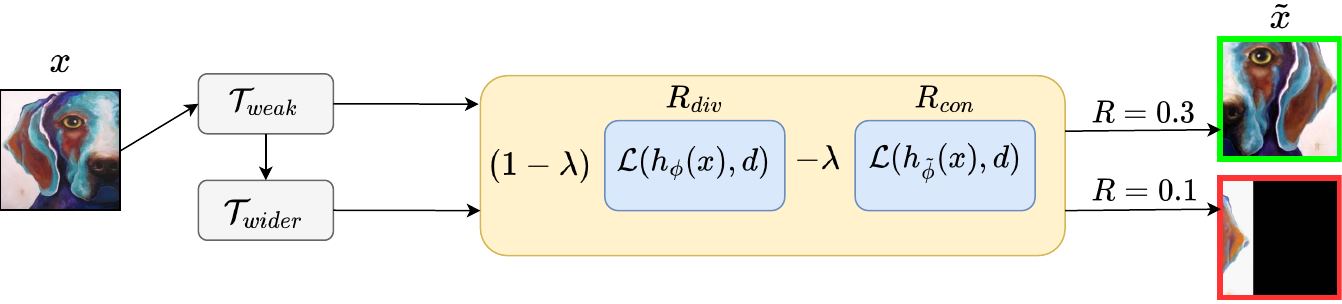}
 
    \caption{Overview of DCAug$^{domain}$ procedure for rejecting extreme augmentations. After calculating $R_{div}$ and $R_{con}$ for $\mathcal{T}_{weak}$ and $\mathcal{T}_{wider}$ our method selects the transformation with the highest reward (green box) and updates the label classifier $f_{\theta}$ and domain student $h_{\phi}$ using the transformed input $\tilde{x}$. DCAug$^{label}$ and TeachDCAug$^{label}$ also follow the same procedure by replacing $d$ and $h_{\phi}$ by $y$ and $f_{\theta}$ respectively (see \ref{fig:visual} from supplementary materials for more visual changes of the selected images).}
    \label{fig:scheme}
 \end{figure*}

\begin{algorithm}[h]
\caption{DCAug Training Procedure}
\label{algo:1}
\begin{algorithmic}[1]
     \renewcommand{\algorithmicrequire}{\textbf{Input:}}
     \renewcommand{\algorithmicensure}{\textbf{Output:}}
     \REQUIRE source domains $S$, label classifier $f_\theta$, domain classifier 
        $h_\phi$, transformations $\mathcal{T}_{weak}$ and $\mathcal{T}_{wider}$, learning rate $\eta$.
        
     \ENSURE  label classifier $f_\theta$ or $f_{\tilde{\theta}}$.
      \FOR {minibatch ${(x,y,d)}$ in training dataset $S$}
      \STATE $\hat{x}_1 \leftarrow \mathcal{T}_{weak} (x)$
      \STATE $\hat{x}_2 \leftarrow \mathcal{T}_{wider} (\hat{x}_1)$
      \STATE select $\hat{x}$ according to Eq. \ref{eq:1}
      \STATE $\theta \leftarrow \theta - \eta \nabla_{\theta}[\mathcal{L}(f_\theta(\hat{x}, y)]$
      \IF {DCAug$^{label}$}
      \STATE $\tilde{\theta} = (1-\beta) \theta + \beta \tilde{\theta}$
      \ELSIF {DCAug$^{domain}$}
      \STATE $\phi \leftarrow \phi - \eta \nabla_{\phi}[\mathcal{L}(h_\phi(\hat{x}, d)]$
      \STATE $\tilde{\phi} = (1-\beta) \phi + \beta \tilde{\phi}$
      \ENDIF
      \ENDFOR
\end{algorithmic} 
\end{algorithm}

To make our idea work, we must ensure that the diversity reward takes into account the latest changes in the model. Thus, as a reward for diversity, we use the cross-entropy loss $\mathcal{L}$ of a classifier $h$ with parameters $\phi$ trained to detect the domain of the image $x$:
\begin{equation}
\begin{split}
R_{div}(x, d) &= \mathcal{L}(h_\phi(x), d)
\end{split}
\label{eq3}
\end{equation}
In this way, the reward would avoid favoring multiple times the same samples because they are already included in the model.
At the same time, the consistency reward needs to be robust because we need to make sure to classify those samples correctly. To do that, we use an exponential moving average (EMA) of our domain classifier $\tilde{\phi}$ as a consistent teacher:
\begin{equation}
\begin{split}
& \tilde{\phi} = (1-\beta) \phi + \beta \tilde{\phi} \\
& R_{con}(x, d) = \mathcal{L}(h_{\tilde{\phi}}(x), d) \\
\end{split}
\label{eq4}
\end{equation}
where $\beta$ defines the smoothness of the moving average and is fixed at $0.999$ for all experiments. We call this approach DCAug$^{domain}$. 

Alternatively, in situations where the domain meta-data $d$ is not available, we can rewrite Eqs. \ref{eq3} and \ref{eq4} by using the label classifier $f_{\theta}$ as the teacher and student, and the class label as ground truth. The method is referred hereafter as DCAug$^{label}$ and uses the rewards terms:
\begin{equation}
\begin{split}
R_{div}(x, y) &= \mathcal{L}(f_\theta(x), y) \\
R_{con}(x, y) &= \mathcal{L}(f_{\tilde{\theta}}(x), y)
\end{split}
\end{equation}
being $\tilde{\theta}$ the exponential moving average (EMA) of $\theta$. This method also gives us the opportunity to use $\tilde{\theta}$ instead of $\theta$ as the final classifier which usually results in a more robust model \cite{arpit2022ensemble}. We call this special variant TeachDCAug$^{label}$. Figure \ref{fig:scheme} shows an overview of the DCAug procedure. For each iteration, our training schema comprises two phases. In the first phase, we freeze $\theta, \phi$ parameters and select the most appropriate transformation based on our reward function $R$. In the second phase, we update $\theta, \phi$ using a gradient descent procedure. The full algorithm of DCAug training procedure is presented in Algorithm \ref{algo:1}. 


\section{Experiments}
\label{sec:5}

\textbf{Dataset.} Following DomainBed benchmark \cite{gulrajani2020search} we evaluate our method on five diverse datasets: \\
PACS \cite{li2017deeper} is a 7-way object classification task with 4 domains and 9,991 samples. VLCS \cite{fang2013unbiased} is a 5-way classification task with 4 domains and 10,729 samples. This dataset mostly contains real photos. The distribution shifts are subtle and simulate real-life scenarios well. OfficeHome \cite{venkateswara2017deep} is a 65-way classification task depicting everyday objects with 4 domains and a total of 15,588 samples. TerraIncognita \cite{beery2018recognition} is a 10-way classification problem of animals in wildlife cameras, where the 4 domains are different locations. There are 24,788 samples. This represents a realistic use case where generalization is indeed critical. DomainNet \cite{peng2019moment} is a 345-way object classification task with 6 domains. With a total of 586,575 samples, DomainNet is larger than most of the other evaluated datasets in both samples and classes. 


\begin{table*}[h]
    \centering
    \begin{tabular}{@{}lccccccc@{}}

        \toprule
        \multirow{2}{*}[-1em]{\textbf{Method}} & \multirow{2}{*}[-1em]{\textbf{Category}} & {} & {} & {} &
        \\
        
        {} & {} & {} & \multicolumn{3}{c}{\multirow{2}{*}[1em]{$\qquad$$\qquad$\textbf{Dataset}}} \\

        \cmidrule(lr){3-8}
        \addlinespace[5pt]

        {} & {} & PACS & VLCS & OfficeHome & TerraInc & DomainNet & \textbf{Avg.}\\
        
        \midrule
        ERM \cite{vapnik1991principles} &
        \emph{Baseline} &
        84.2$\scriptstyle\pm 0.1$  &
        77.3$\scriptstyle\pm 0.1$  &
        67.6$\scriptstyle\pm 0.2$  &
        47.8$\scriptstyle\pm 0.6$  &
        44.0$\scriptstyle\pm 0.1$  &
        64.2  \\

        \hline
        
        MMD \cite{li2018domain} &
        \multirow{4}{*}[-1em]{\emph{Domain-Invariant}} &
        84.7$\scriptstyle\pm 0.5$  &
        77.5$\scriptstyle\pm 0.9$  &
        66.4$\scriptstyle\pm 0.1$  &
        42.2$\scriptstyle\pm 1.6$  &
        23.4$\scriptstyle\pm 9.5$  &
        58.8  \\
        IRM \cite{arjovsky2019invariant} &
        {} &
        83.5$\scriptstyle\pm 0.8$  &
        78.6$\scriptstyle\pm 0.5$  &
        64.3$\scriptstyle\pm 2.2$  &
        47.6$\scriptstyle\pm 0.8$  &
        33.9$\scriptstyle\pm 2.8$  &
        61.6  \\
        GroupDRO \cite{sagawa2019distributionally} &
        {} &
        84.4$\scriptstyle\pm 0.8$  &
        76.7$\scriptstyle\pm 0.6$  &
        66.0$\scriptstyle\pm 0.7$  &
        43.2$\scriptstyle\pm 1.1$  &
        33.3$\scriptstyle\pm 0.2$  &
        60.7  \\
        DANN \cite{ganin2016domain} &
        {} &
        83.7$\scriptstyle\pm 0.4$  &
        78.6$\scriptstyle\pm 0.4$  &
        65.9$\scriptstyle\pm 0.6$  &
        46.7$\scriptstyle\pm 0.5$  &
        38.3$\scriptstyle\pm 0.1$  &
        62.6  \\
        CORAL \cite{sun2016deep} &
        {} &
        86.2$\scriptstyle\pm 0.3$  &
        78.8$\scriptstyle\pm 0.6$  &
        68.7$\scriptstyle\pm 0.3$  &
        47.6$\scriptstyle\pm 1.0$  &
        41.5$\scriptstyle\pm 0.1$  &
        64.5  \\
        mDSDI \cite{bui2021exploiting} &
        {} &
        86.2$\scriptstyle\pm 0.2$  &
        79.0$\scriptstyle\pm 0.3$  &
        69.2$\scriptstyle\pm 0.4$  &
        48.1$\scriptstyle\pm 1.4$  &
        42.8$\scriptstyle\pm 0.2$  &
        65.1  \\
        
        \hline
        DDAIG \cite{zhou2020deep} &
        \multirow{8}{*}[0.5em]{\emph{Data Augmentation}} &
        83.1  &
        -  &
        65.5  &
        -  &
        -  &
        -  \\
        MixStyle \cite{zhou2021domain} &
        {} &
        85.2$\scriptstyle\pm 0.3$  &
        77.9$\scriptstyle\pm 0.5$  &
        60.4$\scriptstyle\pm 0.3$  &
        44.0$\scriptstyle\pm 0.7$  &
        34.0$\scriptstyle\pm 0.1$  &
        60.3  \\
        RSC \cite{huang2020self} &
        {} &
        85.2$\scriptstyle\pm 0.9$  &
        77.1$\scriptstyle\pm 0.5$  &
        65.5$\scriptstyle\pm 0.9$  &
        46.6$\scriptstyle\pm 1.0$  &
        38.9$\scriptstyle\pm 0.5$  &
        62.7 \\
        Mixup \cite{yan2020improve} &
        {} &
        84.6$\scriptstyle\pm 0.6$  &
        77.4$\scriptstyle\pm 0.6$  &
        68.1$\scriptstyle\pm 0.3$  &
        47.9$\scriptstyle\pm 0.8$  &
        39.2$\scriptstyle\pm 0.1$  &
        63.4  \\
        SagNets \cite{nam2021reducing} &
        {} &
        86.3$\scriptstyle\pm 0.2$  &
        77.8$\scriptstyle\pm 0.5$  &
        68.1$\scriptstyle\pm 0.1$  &
        48.6$\scriptstyle\pm 1.0$  &
        40.3$\scriptstyle\pm 0.1$  &
        64.2 \\
        \rowcolor[HTML]{EFEFEF}
        DCAug$^{domain}$ (Ours) &
        {} &
        {86.1}$\scriptstyle\pm 0.9$  &
        {78.9}$\scriptstyle\pm 0.5$  &
        {68.8}$\scriptstyle\pm 0.4$  &
        48.7$\scriptstyle\pm 0.8$  &
        43.7$\scriptstyle\pm 0.3$  &
        65.2 \\
        \rowcolor[HTML]{EFEFEF}
        DCAug$^{label}$ (Ours) &
        {} &
        {86.1}$\scriptstyle\pm 0.7$ &
        78.6$\scriptstyle\pm 0.4$ &
        68.3$\scriptstyle\pm 0.4$ &
        49.3$\scriptstyle\pm 1.5$  &
        {43.8}$\scriptstyle\pm 0.2$  &
        65.2  \\
        \rowcolor[HTML]{EFEFEF}
        TeachDCAug$^{label}$ (Ours) &
        {} &
        88.4$\scriptstyle\pm 0.2$ &
        78.8$\scriptstyle\pm 0.4$ &
        70.4$\scriptstyle\pm 0.2$ &
        51.1$\scriptstyle\pm 1.1$  &
        46.4$\scriptstyle\pm 0.1$  &
        \textbf{67.0}  \\

        \bottomrule
        
    \end{tabular}
\caption{Comparison with domain generalization methods Out-of-domain accuracies on five
domain generalization benchmarks are presented. We highlight the best overall result. For each category, we also report the average accuracy per dataset. Accuracies other than our methods (DCAug) are from \cite{gulrajani2020search, cha2021swad}. Our experiments are repeated three times.}
\label{tab:main_results}
\end{table*}

\textbf{Evaluation protocols and Implementation details.} All performance scores are evaluated by leave-one-out cross-validation, averaging all cases that use a single domain as the target (test) domain and the others as the source (training) domains. We employ DomainBed training and evaluation protocols \cite{gulrajani2020search}. In particular, for training, we use ResNet-50 \cite{he2016deep} pre-trained on the ImageNet
\cite{russakovsky2015imagenet} as default. The model is optimized using Adam \cite{kingma2014adam} optimizer. A mini-batch
contains all domains and 32 examples per domain. For the model hyperparameters, such as learning rate, dropout rate, and weight decay, we use the same configuration as proposed in \cite{cha2021swad}. 
We follow \cite{cha2021swad} and train models for 15000 steps on DomainNet and 5000 steps for other datasets, which corresponds to a variable number of epochs dependent on dataset size. Every experiment is repeated three times with different seeds. We leave 20\% of source domain data for validation. We use training-domain validation for the model selection in which, for each random seed, we choose the model maximizing the accuracy on the validation set. The balancing coefficient of our method, $\lambda$, is coarsely tuned on the validation with three different values: [0.2, 0.5, 0.8].

\subsection{Main Results}

In this section, we compare three variations of our model, DCAug$^{domain}$, DCAug$^{label}$ and TeachDCAug$^{label}$ with and without domain meta-data (as explained in section \ref{sec:4:2}), with 11 related methods in DG.
Those methods are divided into two families: data augmentation and domain-invariant representation. For data augmentation, we compare with Mixup \cite{yan2020improve}, MixStyle \cite{zhou2021domain}, DDAIG \cite{zhou2020deep}, SagNets \cite{nam2021reducing} and RSC \cite{huang2020self}. For invariant representation learning, we compare with IRM \cite{arjovsky2019invariant}, GroupDRO \cite{sagawa2019distributionally}, CORAL \cite{sun2016deep}, MMD \cite{li2018domain}, DANN \cite{ganin2016domain} and mDSDI \cite{bui2021exploiting}. 
We also included ERM as a strong baseline as shown in \cite{gulrajani2020search}.

Table \ref{tab:main_results} shows the overall performance of DCAug and other methods on five domain generalization benchmarks on a classification task. The full result per dataset and the domain are provided in the supplementary material. 
From the table, we observe that, as shown in \cite{idrissi2022simple}, most methods struggle to reach the performance of a simple ERM adapted to multiple domains and only a few methods manage to obtain good results on all datasets. TeachDCAug$^{label}$, which essentially is the moving average version of DCAug$^{label}$, while being simple, manages to rank among the first on all datasets, outperforming all data augmentation and domain-invariant methods. Furthermore, DCAug$^{domain}$ and DCAug$^{label}$ manage to outperform all data augmentation methods and obtain comparable results with the best domain-invariant methods which underlines the importance of good data augmentation for DG.
Another important observation is that as the dataset size increased, from PACS with 9,991 samples to DomainNet with 586,575 samples (see section \ref{sec:5}), most of the methods, especially from the domain-invariant family struggled to show any performance increase compared to the ERM baseline. This poor performance is probably due to the large size and larger number of domains of TerraIncognita and DomainNet which makes all approaches that try to artificially increase the generalization of the algorithm not profitable. On the other hand, our approach manages to keep the performance close or better (in the case of TerraIncognita) to ERM.
Recently, a new method based on ensembling models \cite{rame2022diverse} trained with different hyperparameters has reached an average accuracy of $68$ on the evaluated datasets. However, this approach does not belong to the studied methods and is orthogonal to them. 
Compared to ERM, DCAug has a small additional computational cost. In particular, DCAug$^{domain}$, other than updating the parameters of both domain and label classifier for each sample, computes the loss of the domain classifier twice without the need to calculate the gradients (see \ref{sec:compute} from supplementary materials for full characterization.)

\subsection{Empirical analysis of DCAug}

\begin{figure}[h]
   \centering
    \includegraphics[width=\linewidth]{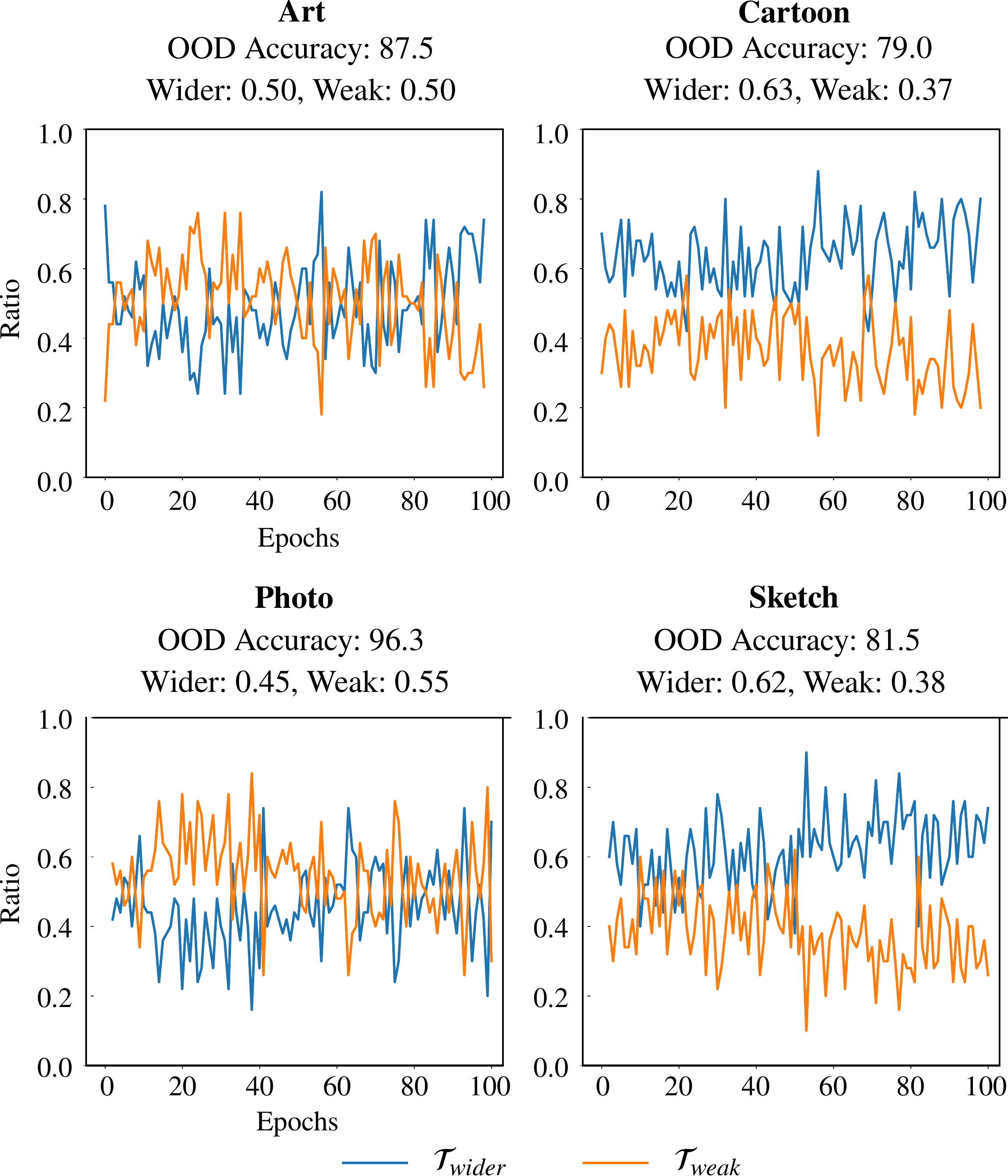}
 
    \caption{Evolution over epochs of Weak vs Wider augmentations on the four domains of the PACS dataset. The title of each plot shows the domain, out-of-domain accuracies, and the average ratio of each augmentation for the entire training run. In the plots, we clearly see that for Cartoon and Sketch, the two domains that are farther from the pre-trained model on ImageNet and with lower performance, strong transformations are preferred over weak ones.}
    \label{fig:abl_2}
 \end{figure}

\textbf{Rejection rate of strong augmentations.} We study the evolution of the rejection rate of strong augmentations over epochs on the PACS dataset. We use the same balancing hyperparameter $\lambda$ as in the main experiments. For each domain, we show the selection of weak and strong augmentations for the entire training. As we observe in Figure \ref{fig:abl_2}, the rejecting rate is domain-dependent. As models have been pre-trained on ImageNet \cite{russakovsky2015imagenet}, we can see that our method selects weak and strong augmentations equally for domains that are closer to the pre-trained dataset (Art and Photo). However, for Cartoon and Sketch, which are far from ImageNet, we observe that our method uses more strong augmentations. This observation is in line with other works that suggest the effect of data augmentations diminishes if the training data already covers most of the variations in the dataset \cite{wiles2021fine, liu2022empirical}. 

\begin{figure*}[h]
   \centering
    \includegraphics[width=0.9\linewidth]{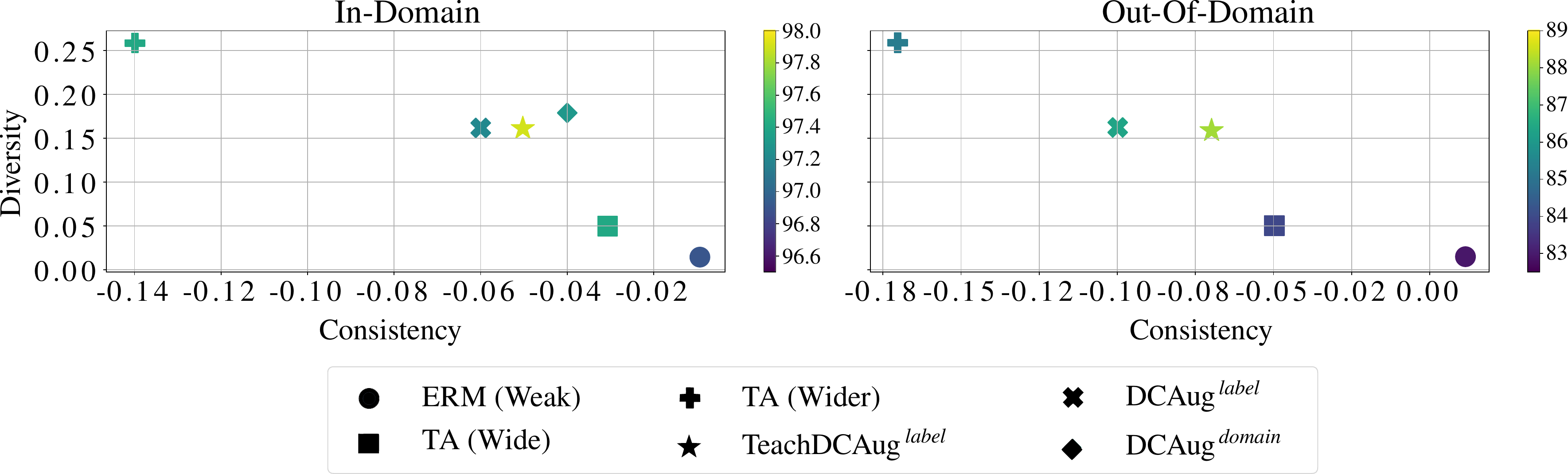}
 
    \caption{Consistency and diversity for different methods for in-domain (left) and out-of-domain (right) settings on the PACS dataset. Color represents the classification accuracy on the test set. 
    For high accuracy, we need a good trade-off between diversity and consistency.}
    \label{fig:abl_affinity}
\end{figure*}

\textbf{Measuring diversity and consistency.} Following the \cite{gontijo2020affinity}, in this section we compare the diversity and consistency (affinity) of the transformation space that our proposed approach induces with TA (Wide), TA (Wider), and ERM. Intuitively consistency measures the level of distortion caused by a given data augmentation schema on the target dataset. Here in our case, we use the in-domain validation and out-domain test set to measure in and out-domain performance respectively. On the other hand, diversity is a model-dependent element and captures the difficulty of the model to fit the augmented training data (see \cref{supp-sec:affinity} from supplementary materials for precise definitions). As shown in Figure \ref{fig:abl_affinity}, neither of the two extremes, TA (Wider) as the most diverse and ERM with standard data augmentations as the most consistent, is sufficient for the best final performance. However, all of our proposed approaches provide a good trade-off between consistency and diversity which results in the best final performance for both in-domain and out-domain.

\begin{table*}[h]
    \centering
    \begin{tabular}{@{}lccccccc@{}}

        \toprule
        \multirow{2}{*}[-1em]{\textbf{Method}} & \multirow{2}{*}[-1em]{\textbf{Search Space}} & {} & {} & {} &
        \\
        
        {} & {} & {} & \multicolumn{3}{c}{\multirow{2}{*}[1em]{$\qquad$$\qquad$\textbf{Dataset}}} \\

        \cmidrule(lr){3-8}
        \addlinespace[5pt]

        {} & {} & PACS & VLCS & OfficeHome & TerraInc & DomainNet & \textbf{Avg.}\\
        
        \midrule

        AutoAugment \cite{cubuk2019autoaugment} &
        {Default} &
        97.3$\scriptstyle\pm 0.2$  &
        87.0$\scriptstyle\pm 0.1$  &
        82.9$\scriptstyle\pm 0.3$  &
        90.1$\scriptstyle\pm 0.2$  &
        62.2$\scriptstyle\pm 0.1$  &
        83.9 \\

        TA (Ours) &
        {Wider} &
        97.6$\scriptstyle\pm 0.1$  &
        87.2$\scriptstyle\pm 0.1$  &
        83.4$\scriptstyle\pm 0.3$  &
        89.7$\scriptstyle\pm 0.1$  &
        62.0$\scriptstyle\pm 0.1$  &
        84.0 \\

        RandAugment \cite{cubuk2020randaugment} &
        {Default} &
        97.3$\scriptstyle\pm 0.2$  &
        87.1$\scriptstyle\pm 0.2$   &
        82.8$\scriptstyle\pm 0.3$   &
        91.2$\scriptstyle\pm 0.1$  &
        62.4$\scriptstyle\pm 0.4$   &
        84.2 \\

        TA \cite{muller2021trivialaugment} &
        {Wide} &
        97.6$\scriptstyle\pm 0.2$  &
        87.2$\scriptstyle\pm 0.1$  &
        83.8$\scriptstyle\pm 0.4$  &
        90.9$\scriptstyle\pm 0.2$  &
        62.7$\scriptstyle\pm 0.1$  &
        84.4 \\
        
        DCAug$^{domain}$ (Ours) &
        {Wider} &
        97.5$\scriptstyle\pm 0.2$  &
        87.1$\scriptstyle\pm 0.1$  &
        83.5$\scriptstyle\pm 0.3$  &
        91.6$\scriptstyle\pm 0.2$  &
        62.8$\scriptstyle\pm 0.1$  &
        84.5 \\

        DCAug$^{label}$ (Ours) &
        {Wider} &
        97.6$\scriptstyle\pm 0.2$  &
        87.3$\scriptstyle\pm 0.2$  &
        83.4$\scriptstyle\pm 0.4$  &
        91.3$\scriptstyle\pm 0.2$  &
        62.8$\scriptstyle\pm 0.1$  &
        84.5 \\

        TeachDCAug$^{label}$ (Ours) &
        {Wider} &
        98.1$\scriptstyle\pm 0.2$  &
        87.5$\scriptstyle\pm 0.2$  &
        84.1$\scriptstyle\pm 0.4$  &
        92.5$\scriptstyle\pm 0.1$  &
        65.6$\scriptstyle\pm 0.1$  &
        \textbf{85.6} \\

        \bottomrule
    \end{tabular}
\caption{In-domain accuracies of our methods on five domain generalization benchmarks. Our experiments are repeated three times.}
\label{tab:in_domain}
\end{table*}

\textbf{In-domain performance.} We investigate the in-domain performance of Autoaugment, RandAugment, TA (Wide), TA (Wider), and our proposed methods in Table \ref{tab:in_domain}. As we can see it has the exact opposite ranking compared to Table \ref{tab:motivation} which shows that aggressive data augmentation can indeed harm the in-domain performance. Our method, however, has proven to be highly effective in both cases, demonstrating that by limiting the scope of transformation, we can achieve optimal outcomes that combine the best of both worlds.

\begin{table}[h]
    \centering
    \begin{tabular}{lcc}
        \toprule
        \textbf{Diversity} & \textbf{Consistency} & \textbf{OOD Accuracy} \\
        \addlinespace[5pt]
        \midrule
        $\mathcal{L}(h_\phi(x), d)$ & $\mathcal{L}(f_\theta(x), y)$ & 81.8 \\
        $\mathcal{L}(h_{\tilde\phi}(x), d)$ & $\mathcal{L}(f_{\tilde\theta}(x), y)$ & 80.1 \\
        $\mathcal{L}(f_\theta(x), y)$ & $\mathcal{L}(f_{\tilde\theta}(x), y)$ & \textbf{86.1} \\
        $\mathcal{L}(h_\phi(x), d)$ & $\mathcal{L}(h_{\tilde\phi}(x), d)$ & \textbf{86.1} \\
        \bottomrule
    \end{tabular}
\caption{Variations of our reward function on {PACS dataset}. $h_{\tilde\phi}$ and $f_{\tilde\theta}$ refer to the EMA version of their corresponding models. Our methods are highlighted in the table.}
\label{tab:abl_2}
\end{table}

\textbf{Diverse domain and consistent label.} Given that our final target is to select a transformation that creates challenging samples with diverse domains without losing the sample’s original meaning, one might want to use the domain classifier and label classifier to satisfy this goal. In particular, we can write a variation of our reward functions as follows:
\begin{equation}
\begin{split}
R_{div}(x, d) &= \mathcal{L}(h_\phi(x), d) \\
R_{con}(x, y) &= \mathcal{L}(f_\theta(x), y),
\end{split}
\end{equation}
in which the diversity is measured in terms of domain labels $d$ and the consistency in terms of class labels $y$.
This formulation resembles the reward function used in \cite{zhou2020deep}, although for different purposes. We also derive another variation to this reward function which uses the EMA version of each model. Also for these rewards, we tune the hyperparameter $\lambda$ to find the best balance between diversity and consistency. As reported in Table \ref{tab:abl_2}, these two variants of our formulation do not really work. In particular, since the task of domain classification is significantly easier than the target task, finding a good balance between these two turns out to be difficult \cite{setlur2023bitrate}

\section{Conclusion} \label{sec:6}

In this work, we have presented a method for improving the performance of data augmentation when multiple domains are available at training time, but the distribution of the test domain is different from those and unknown.
In this setting, we show that the state-of-the-art in-domain augmentation of TrivialAugment \cite{muller2021trivialaugment} based on uniform sampling of predefined transformation is beneficial and helps to improve results for a baseline based on ERM, which has been shown to be strong for domain generalization \cite{gulrajani2020search}. 
Then, we propose to further improve results by increasing the magnitude of those transformations while keeping the random sampling. This makes sense for domain generalization as we want to learn an unknown domain. 
Finally, we propose a rejection scheme that removes extreme and harmful transformations during training based on a reward function that compares the performance of a label classifier with an exponential moving average of it. All these contributions allowed our method to achieve equal or better results than state-of-the-art methods on five challenging domain generalization datasets with a minimum intervention in the standard ERM pipeline.

{\small
\bibliographystyle{ieee_fullname}
\bibliography{main}
}

\begin{appendices}

In this supplementary material, we give additional information to reproduce our work. Here, we provide implementation details, characterization of computational cost, the definition of affinity and diversity, visual changes of the selected by our proposed data augmentation schema, the effect of ViT-backbone, the effect of hyperparameter $\lambda$, and finally, show detailed results of Table 3 in the main manuscript.

\section{Implementation details}


The evaluation protocol by \cite{gulrajani2020search} is computationally too expensive, therefore we use the reduced search space from \cite{cha2021swad} for the common parameters. Table \ref{tab:hyp} summarizes the hyperparameter search space. We use the same search space for all datasets. To further reduce the hyperparameter search, we start by finding the optimal hyperparameter for TA and then use those to find the best $\lambda$ for our proposed method.

\begin{table}[h]
    \centering
    \begin{tabular}{lc}
        \toprule
        \textbf{Hyperparameter} & \textbf{Search Space} \\
        \addlinespace[1pt]
        \midrule
        batch size &
        32 \\
        learning rate &
        \{1e-5, 3e-5, 5e-5\} \\
        ResNet dropout &
        \{0.0, 0.1, 0.5\} \\
        weight decay &
        \{1e-4, 1e-6\} \\
        \hline
        \addlinespace[1pt]
        $\lambda$ &
        \{0.2, 0.5, 0.8\} \\
        \bottomrule
    \end{tabular}
  \caption{Hyperparameters used for all methods in and their respective distributions for grid search. $\lambda$ refers to the balancing coefficient of the proposed reward function (Eq 2. of the manuscript).}
  \label{tab:hyp}
\end{table}

\subsection{Datasets}

\textbf{PACS:} \cite{li2017deeper} is a 7-way object classification task with 4 domains: art, cartoon, photo, and sketch, with 9,991 samples. 

\textbf{VLCS:} \cite{fang2013unbiased} is a 5-way classification task from 4 domains: Caltech101, LabelMe, SUN09, and VOC2007. There are 10,729 samples. This dataset mostly contains real photos. The distribution shifts are subtle and simulate real-life scenarios well. 

\textbf{OfficeHome:} \cite{venkateswara2017deep} is a 65-way classification task depicting everyday objects from 4 domains: art, clipart, product, and real, with a total of 15,588 samples. 

\textbf{TerraIncognita:} \cite{beery2018recognition} is a 10-way classification problem of animals in wildlife cameras, where the 4 domains are different locations, L100, L38, L43, L46. There are 24,788 samples. This represents a realistic use case where generalization is indeed critical. 

\textbf{DomainNet:} \cite{peng2019moment} is a 345-way object classification task from 6 domains: clipart, infograph, painting, quickdraw, real, and sketch. With a total of 586,575 samples, it is larger than most of the other evaluated datasets in both samples and classes. 

\subsection{Code}

Our work is built upon DomainBed\footnote{\url{https://github.com/facebookresearch/DomainBed}} \cite{gulrajani2020search} and SWAD\footnote{\url{https://github.com/khanrc/swad}} \cite{cha2021swad} codebase, which is released under the MIT license.

\begin{table*}[h]
    \centering
    \begin{tabular}{@{}lccc@{}}
      
      \toprule

        \multirow{2}{*}[-1em]{\textbf{Transform}}  & {} \\
        
        {} &\multicolumn{3}{c}{\multirow{2}{*}[1em]{\textbf{Search Space Ranges} }} \\

        \cmidrule(lr){2-4}
        \addlinespace[5pt]

        {}  & \textbf{Default \cite{muller2021trivialaugment, cubuk2020randaugment}} &  \textbf{Wide \cite{muller2021trivialaugment}} & \textbf{Wider (Ours)} \\

        \midrule
        ShearX(Y) & [-0.3, 0.3] & [-1.0, 1.0] & [-1.0, 1.0] \\ [0.5ex] 
        TranslateX(Y) & [-32, 32] & [-32, 32] & [-224.0, 224.0] \\ [0.5ex] 
        Rotate & [-30.0, 30.0] & [-135.0, 135.0] & [-135.0, 135.0] \\ [0.5ex] 
        Posterize & [4, 8] & [2, 8] & [0, 8] \\ [0.5ex] 
        Solarize & [0, 255] & [0, 255] & [0, 255]  \\ [0.5ex] 
        Contrast & [-1.0, 1.0] & [-1.0, 1.0] & [-10.0, 10.0] \\ [0.5ex]  
        Color & [-1.0, 1.0] & [-1.0, 1.0] & [-10.0, 10.0] \\ [0.5ex]  
        Sharpness & [-1.0, 1.0] & [-1.0, 1.0] & [-10.0, 10.0] \\ [0.5ex]  
        Brightness & [-1.0, 1.0] & [-1.0, 1.0] & [-1.0, 10.0]  \\ [0.5ex]  
        AutoContrast & N/A & N/A & N/A  \\ [0.5ex]  
        Equalize & N/A & N/A & N/A  \\ [0.5ex]
        Grey & N/A & N/A & N/A  \\ 
        \bottomrule
        
    \end{tabular}
    \caption{List of image transformations and their search space ranges. The table shows the Default range from RandAugment \cite{cubuk2020randaugment}, the Wide range from TA \cite{muller2021trivialaugment} as well as our proposed Wider range.}
	\label{tab:space}
\end{table*}
    
\section{Effect of the balancing coefficient $\lambda$.}

\begin{figure}[t]
  \centering
   \def\svgwidth{\linewidth}
    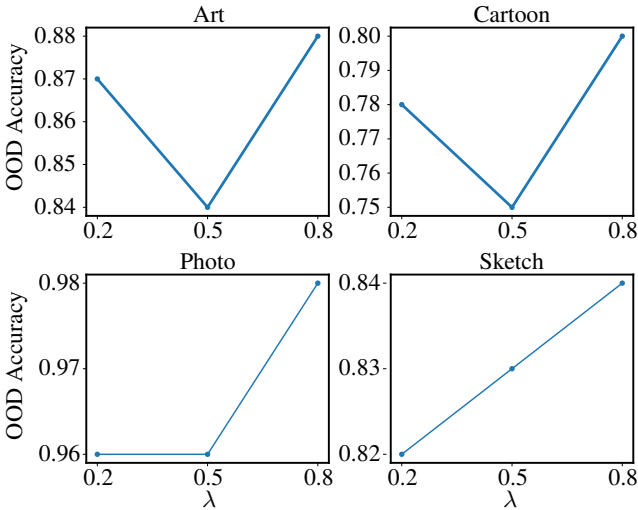
   \caption{Effect of hyperparameter $\lambda$ on PACS dataset. Each plot represents one domain: Art, Cartoon, Photo and Sketch. On x-axis is the $\lambda$ and on y-axis the OOD accuracy.}
   \label{fig:pacs}
\end{figure}

\begin{figure}[b]
    \centering
    \def\svgwidth{\linewidth}
    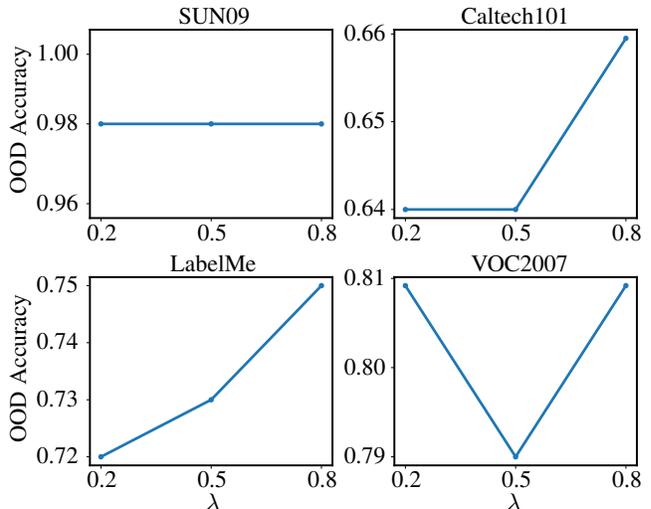
    \caption{Effect of hyperparameter $\lambda$ on VLCS dataset. Each plot represents one domain: SUN09, Caltech101, LabelMe, and VOC2007. On the x-axis is the $\lambda$ and on the y-axis is the OOD accuracy.}
    \label{fig:vlcs}
\end{figure}

\begin{figure}[t]
    \centering
    \def\svgwidth{\linewidth}
    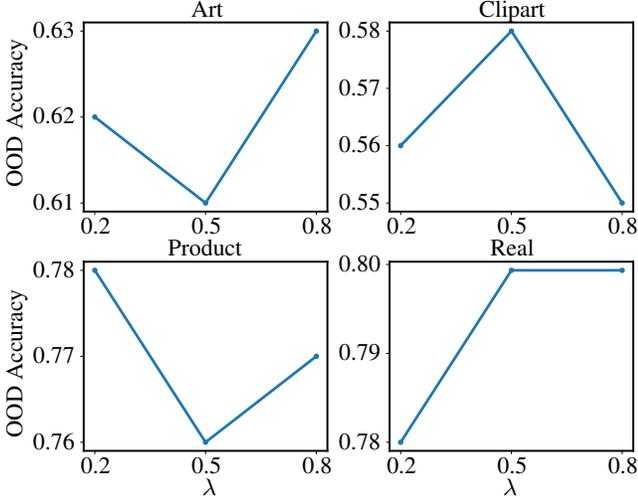
    \caption{Effect of hyperparameter $\lambda$ on OfficeHome dataset. Each plot represents one domain: Art, Clipart, Product, and Real. On the x-axis is the $\lambda$ and on the y-axis is the OOD accuracy.}
    \label{fig:office}
\end{figure}

\begin{figure}[b]
    \centering
    \def\svgwidth{\linewidth}
    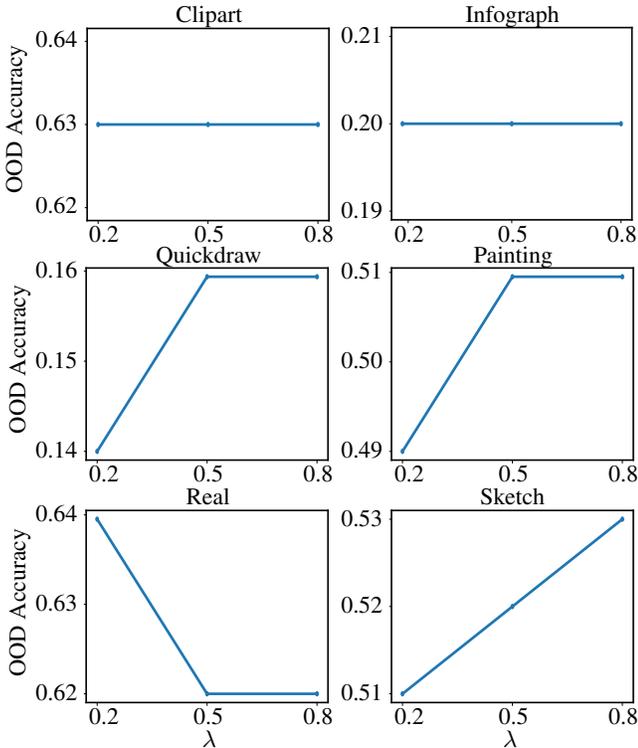
    \caption{Effect of hyperparameter $\lambda$ on DomainNet dataset. Each plot represents one domain: Clipart, Infograph, Painting, Quickdraw, Real, and Sketch. On the x-axis is the $\lambda$ and on the y-axis is the OOD accuracy.}
    \label{fig:domain}
\end{figure}

Figures \ref{fig:pacs}, \ref{fig:vlcs}, \ref{fig:office} and \ref{fig:domain} show the effect of the balancing coefficient between diversity and consistency rewards terms, $\lambda$, on PACS, VLCS, OfficeHome and DomainNet datasets respectively. Figure~\ref{fig:pacs} shows the obtained OOD accuracy for the PACS dataset. The best value of $\lambda$ inside our searching space for all domains was $0.8$. Thus the consistency value was $0.8$ and $0.2$ for the diversity value, which implies that for an improvement on the OOD accuracy, for the PACS, we should go towards a higher value of $\lambda$ for the term of consistency than diversity. This has a positive impact on the performance with gains more than $0.01$ for higher values of $\lambda$, when compared with smaller values for the Photo domain, i.e., for $\lambda=0.2$ the OOD acc. is $0.96$ and for $\lambda=0.5$ the OOD acc. is $0.97$. In which the best value of OOD acc. was $0.98$ for the $\lambda=0.8$ for this domain.

Regarding the Art and Cartoon domains, the extreme cases seem to be better than being too conservative with a $\lambda$ of $0.5$, so if it goes for the extremes $\lambda=0.2$ or $\lambda=0.8$, it is better, with better results for $0.8$. In the Cartoon domain, the $0.8$ $\lambda$ had $0.02$ gain in OOD acc. compared with performance for $\lambda=0.2$, and $0.05$ gain compared with the $\lambda=0.5$. In the case of the Art domain, the previous behavior remained the same where  $\lambda=0.8$ had a gain of $0.01$ when compared with the performance of $\lambda=0.2$ and $0.04$ gain when compared with $\lambda=0.5$. 

Considering Figure~\ref{fig:vlcs}, the best $\lambda$ inside our searching space for all domains was again $0.8$, i.e., the consistency has the importance of $0.8$ against the $0.2$ for the diversity value. Robustness to the chosen $\lambda$ was observed for the SUN09 domain. In Caltech101, $\lambda=0.8$ had a gain of $0.01$ compared with the other two. Similarly, for LabelMe domain the $\lambda=0.8$ was the best, with improvement of $0.03$ and $0.02$ for $\lambda=0.2$ and $\lambda=0.5$, respectively. So for this domain, a high $\lambda$ for consistency value is better than a lower one. For the VOC2007, both extreme cases are good, but $\lambda=0.5$ was worse by $0.01$ compared with $\lambda=0.2$ and $\lambda=0.8$. The OOD accuracy obtained in both cases was $0.80$.

As shown in Figure~\ref{fig:office}, the domain Art had $0.01$ gain with $\lambda=0.8$ when compared with $\lambda=0.2$  and $0.02$ gain compared with $\lambda=0.5$, so far for this domain art, the consistency improved more than the diversity. Considering domain Clipart, the $\lambda=0.5$ was the best with $0.58$ acc. OOD, so consistency and diversity are important for this domain, which has $0.02$ gain compared with $\lambda=0.2$ and $0.03$ gain compared with $\lambda=0.8$. For domain Product, the $\lambda=0.2$ had $0.78$ OOD acc., and it was better than $\lambda=0.5$ and $\lambda=0.8$ by $0.01$ and $0.02$ respectively. For the Real domain, it required a balance of consistency and diversity, or higher $\lambda$, i.e., greater than $0.5$, to gain on the OOD acc.. The best performance obtained was of $0.80$.

As illustrated in Figure~\ref{fig:domain}, the Clipart and Infograph domains were not impacted by the value of $\lambda$ in terms of OOD accuracy. On the other hand, in the Painting domain, a $\lambda$ greater than $0.5$ is preferable, with an increase on the OOD acc. of $0.01$ for $\lambda=0.5$ when compared with $\lambda=0.2$. The same trend occurred with Quickdraw domain with $0.15$ OOD acc. Regarding the Real domain, more diversity can increase the OOD acc., so the $\lambda=0.2$ was better than values of $0.5$ and $0.8$. For such a value of  $\lambda$ the performance was $0.63$. Finally, in the Sketch domain, the performance was linearly correlated with the value of $\lambda$. So $\lambda=0.8$ had the best OOD acc. with $0.53$ value, which represents an increase of $0.02$ compared with $\lambda=0.2$ and $0.01$ increase compared with $\lambda=0.5$.

\section{Computational cost} \label{sec:compute}

\begin{table}[h]
    \centering
    \begin{tabular}{lc}
        \toprule
        \textbf{Method} & \textbf{Minibatch Time (s)} \\
        \addlinespace[5pt]
        \midrule
        ERM & 0.13 \\
        DCAug$^{domain}$ & 0.25 \\
        DCAug$^{label}$ & 0.21 \\
        TeachDCAug$^{label}$ & 0.21 \\
        \bottomrule
    \end{tabular}
\caption{Training iteration time for a minibatch of 32 samples on {PACS dataset} for ERM and our methods.}
\label{tab:compute}
\end{table}

Compared to ERM, DCAug has a small additional computational cost. In particular, DCAug$^{domain}$, other than updating the parameters of both domain and label classifier, for each sample computes the loss of the domain classifier twice without the need to calculate the gradients. As we can the in Table \ref{tab:compute}, on an NVIDIA-A100 GPU, this roughly amounts to twice the slower step time than regular ERM.

\section{Quantifying mechanisms of data Augmentation using affinity and diversity} \label{sec:affinity}

We use the following definition of Affinity and Diversity (as defined in \cite{gontijo2020affinity}): \\

\textbf{Affinity (Consistency):} \emph{Let $a$ be an augmentation and $D_{train}$ and $D_{val}$ be training and validation datasets drawn IID from the same clean data distribution, and let $D'_{val}$ be derived from $D_{val}$ by applying a stochastic augmentation strategy, $a$, once to each image in $D_{val}, D'_{val} = \{(a(x), y): \forall (x, y) \in D_{val} \} $. Further let $m$ be a model trained on $D_{train}$ and $A(m, D)$ denote the model’s accuracy when evaluated on dataset $D$. The Affinity, $T[a; m; D_{val}]$, is given by}

\begin{equation}
T[a; m; D_{val}] = A(m, D_{val}) - A(m, D'_{val}).
\end{equation}

\textbf{Diversity:} \emph{Let $a$ be an augmentation and $D'_{train}$ be the augmented training data resulting from applying the augmentation, $a$, stochastically. Further, let $L_{train}$ be the training loss for a model, $m$,
trained on $D'_{train}$. We define the Diversity, $D[a; m; D_{train}]$, as
}

\begin{equation}
D[a; m; D_{train}] = \mathbb{E}_{D'_{train}} [L_{train}].
\end{equation}

\section{DCAug with ViT backbone} 

In this section, we investigate the robustness of the proposed method to the choice of pretrained models, particularly the ViT backbone \cite{dosovitskiy2020image}. To be able to compare with Resnet50, we use the ViT-B-16 variant which is the base model with a patch size of 16. We use the same experimental setup as before and use the PACS dataset to evaluate the models. As we can see from Table \ref{tab:vit}, while TA seems to lose accuracy when using ViT, our approach shows consistent improvements over the ERM baseline. 

\begin{table}[h]
    \centering
    \begin{tabular}{lc}
        \toprule
        \textbf{Method} & \textbf{OOD Accuracy} \\
        \addlinespace[5pt]
        \midrule
        ERM & 85.0$\scriptstyle\pm 1.1$ \\
        TA & 83.8$\scriptstyle\pm 1.0$ \\
        DCAug$^{domain}$ & 85.4$\scriptstyle\pm 0.6$ \\
        DCAug$^{label}$ & 84.3$\scriptstyle\pm 0.9$ \\
        TeachDCAug$^{label}$ & 88.4$\scriptstyle\pm 0.5$ \\
        \bottomrule
    \end{tabular}
\caption{Out-of-domain performance of models based on ViT-B-16 backbone on PACS dataset. Our experiments are repeated three times.}
\label{tab:vit}
\end{table}

\section{Full Results}

In this section, we show detailed results of Table 3 of the main manuscript. Tables \ref{tab:pacs}, \ref{tab:vlcs}, \ref{tab:office}, \ref{tab:terra} \ref{tab:domainnet} show full results on PACS, VLCS, OfficeHome, TerraIncognita and DomainNet datasets, respectively. The provided tables summarize the obtained out-of-distribution accuracy for every domain within the four datasets. Standard deviations are reported from three trials, when available. The results for the methods were gathered from \cite{gulrajani2020search} and \cite{cha2021swad}. To guarantee the comparability of the results, we followed the same experimental setting as in DomainBed \cite{gulrajani2020search}.

\begin{figure*}[t]
   \centering
    \includegraphics[width=\linewidth]{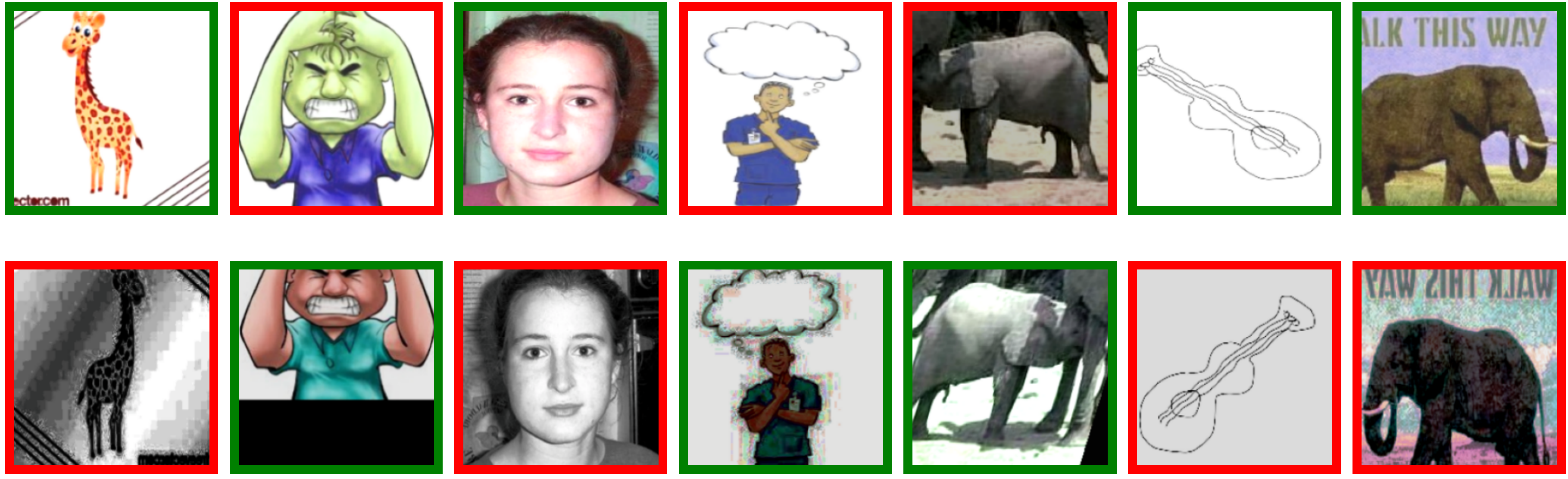}
 
    \caption{Visual changes of the selected images by our proposed data augmentation schema. For each sample within a minibatch, our method produces two augmentations of $\mathcal{T}_{weak}$ (top row) and $\mathcal{T}_{wider}$ (bottom row). After calculating $R_{div}$ and $R_{con}$ for each, our method selects the transformation with the highest reward (green box) and rejects the other one (red box).} 
    \label{fig:visual}
 \end{figure*}

\begin{table*}[h]
    \centering
    \begin{tabular}{@{}lcccccc@{}}

        \toprule
        \multirow{2}{*}[-1em]{\textbf{Method}} & \multirow{2}{*}[-1em]{\textbf{Category}} & {} & {} & {} &
        \\
        
        {} & {} & {} & \multicolumn{2}{c}{\multirow{2}{*}[1em]{$\qquad$$\qquad$\textbf{Domain}}} \\

        \cmidrule(lr){3-7}
        \addlinespace[5pt]

        {} & {} & Art & Cartoon & Photo & Sketch & \textbf{Avg.}\\
        
        \midrule
        ERM \cite{vapnik1991principles} &
        \emph{Baseline} &
        85.7$\scriptstyle\pm 0.6$  &
        77.1$\scriptstyle\pm 0.8$  &
        97.4$\scriptstyle\pm 0.4$  &
        76.6$\scriptstyle\pm 0.7$  &
        84.2 \\
        \hline

        MMD \cite{li2018domain} &
        \multirow{4}{*}[-1em]{\emph{Domain-Invariant}} &
        86.1$\scriptstyle\pm 1.4$  &
        79.4$\scriptstyle\pm 0.9$  &
        96.6$\scriptstyle\pm 0.2$  &
        76.5$\scriptstyle\pm 0.5$  &
        84.7  \\
        IRM \cite{arjovsky2019invariant} &
        {} &
        84.8$\scriptstyle\pm 1.3$  &
        76.4$\scriptstyle\pm 1.1$  &
        96.7$\scriptstyle\pm 0.6$  &
        76.1$\scriptstyle\pm 1.0$  &
        83.5  \\
        GroupDRO \cite{sagawa2019distributionally} &
        {} &
        83.5$\scriptstyle\pm 0.9$  &
        79.1$\scriptstyle\pm 0.6$  &
        96.7$\scriptstyle\pm 0.3$  &
        78.3$\scriptstyle\pm 2.0$  &
        84.4  \\
        DANN \cite{ganin2016domain} &
        {} &
        86.4$\scriptstyle\pm 0.8$  &
        77.4$\scriptstyle\pm 0.8$  &
        97.3$\scriptstyle\pm 0.4$  &
        73.5$\scriptstyle\pm 2.3$  &
        83.7  \\
        CORAL \cite{sun2016deep} &
        {} &
        88.3$\scriptstyle\pm 0.2$  &
        80.0$\scriptstyle\pm 0.5$  &
        97.5$\scriptstyle\pm 0.3$  &
        78.8$\scriptstyle\pm 1.3$  &
        86.2  \\
        mDSDI \cite{bui2021exploiting} &
        {} &
        87.7$\scriptstyle\pm 0.4$  &
        80.4$\scriptstyle\pm 0.7$  &
        98.1$\scriptstyle\pm 0.3$  &
        78.4$\scriptstyle\pm 1.2$  &
        86.2  \\
        \hline
       DDAIG \cite{zhou2020deep} &
       \multirow{8}{*}[0.5em]{\emph{Data Augmentation}} &
       84.2  &
       78.1  &
       95.3  &
       74.7  &
       83.1  \\
       MixStyle \cite{zhou2021domain} &
       {} &
       86.8$\scriptstyle\pm 0.5$  &
       79.0$\scriptstyle\pm 1.4$  &
       96.6$\scriptstyle\pm 0.1$  &
       78.5$\scriptstyle\pm 2.3$  &
       85.2  \\
       RSC \cite{huang2020self} &
       {} &
       85.4$\scriptstyle\pm 0.8$  &
       79.7$\scriptstyle\pm 1.8$  &
       97.6$\scriptstyle\pm 0.3$  &
       78.2$\scriptstyle\pm 1.2$  &
       85.2 \\
       Mixup \cite{yan2020improve} &
       {} &
       86.1$\scriptstyle\pm 0.5$  &
       78.9$\scriptstyle\pm 0.8$  &
       97.6$\scriptstyle\pm 0.1$  &
       75.8$\scriptstyle\pm 1.8$  &
       84.6  \\
       SagNets \cite{nam2021reducing} &
       {} &
       87.4$\scriptstyle\pm 1.0$  &
       80.7$\scriptstyle\pm 0.6$  &
       97.1$\scriptstyle\pm 0.1$  &
       80.0$\scriptstyle\pm 0.4$  &
       86.3 \\
       \rowcolor[HTML]{EFEFEF}
       DCAug$^{domain}$ (Ours) &
       {} &
       87.5$\scriptstyle\pm 0.7$  &
       79.0$\scriptstyle\pm 1.5$  &
       96.3$\scriptstyle\pm 0.1$  &
       81.5$\scriptstyle\pm 0.9$  &
       86.1 \\
       \rowcolor[HTML]{EFEFEF}
       DCAug$^{label}$ (Ours) &
       {} &
       88.5$\scriptstyle\pm 0.8$ &
       78.8$\scriptstyle\pm 1.5$ &
       96.3$\scriptstyle\pm 0.1$ &
       80.8$\scriptstyle\pm 0.5$  &
       86.1  \\
       \rowcolor[HTML]{EFEFEF}
       TeachDCAug$^{label}$ (Ours) &
       {} &
       89.6$\scriptstyle\pm 0.0$ &
       81.8$\scriptstyle\pm 0.5$ &
       97.7$\scriptstyle\pm 0.0$ &
       84.5$\scriptstyle\pm 0.2$  &
       88.4  \\

        \bottomrule
    \end{tabular}
  \caption{Out-of-domain accuracies (\%) on PACS.}
   \label{tab:pacs}
\end{table*}

\begin{table*}[h]
    \centering
    \begin{tabular}{@{}lcccccc@{}}

        \toprule
        \multirow{2}{*}[-1em]{\textbf{Method}} & \multirow{2}{*}[-1em]{\textbf{Category}} & {} & {} & {} &
        \\
        
        {} & {} & {} & \multicolumn{2}{c}{\multirow{2}{*}[1em]{$\qquad$$\qquad$\textbf{Domain}}} \\

        \cmidrule(lr){3-7}
        \addlinespace[5pt]

        {} & {} & Caltech101 & LabelMe & SUN09 & VOC2007 & \textbf{Avg.}\\
        
        \midrule
        ERM \cite{vapnik1991principles} &
        \emph{Baseline} &
        98.0$\scriptstyle\pm 0.3$  &
        64.7$\scriptstyle\pm 1.2$  &
        71.4$\scriptstyle\pm 1.2$  &
        75.2$\scriptstyle\pm 1.6$  &
        77.3 \\
        \hline
        
        MMD \cite{li2018domain} &
        \multirow{4}{*}[-1em]{\emph{Domain-Invariant}} &
       97.7$\scriptstyle\pm 0.1$  &
       64.0$\scriptstyle\pm 1.1$  &
       72.8$\scriptstyle\pm 0.2$  &
       75.3$\scriptstyle\pm 3.3$  &
       77.5  \\
       IRM \cite{arjovsky2019invariant} &
       {} &
       98.6$\scriptstyle\pm 0.1$  &
       64.9$\scriptstyle\pm 0.9$  &
       73.4$\scriptstyle\pm 0.6$  &
       77.3$\scriptstyle\pm 0.9$  &
       78.6  \\
       GroupDRO \cite{sagawa2019distributionally} &
       {} &
       97.3$\scriptstyle\pm 0.3$  &
       63.4$\scriptstyle\pm 0.9$  &
       69.5$\scriptstyle\pm 0.8$  &
       76.7$\scriptstyle\pm 0.7$  &
       76.7  \\
       DANN \cite{ganin2016domain} &
       {} &
       99.0$\scriptstyle\pm 0.3$  &
       65.1$\scriptstyle\pm 1.4$  &
       73.1$\scriptstyle\pm 0.3$  &
       77.2$\scriptstyle\pm 0.6$  &
       78.6  \\
       CORAL \cite{sun2016deep} &
       {} &
       98.3$\scriptstyle\pm 0.1$  &
       66.1$\scriptstyle\pm 1.2$  &
       73.4$\scriptstyle\pm 0.3$  &
       77.5$\scriptstyle\pm 1.2$  &
       78.8  \\
       mDSDI \cite{bui2021exploiting} &
       {} &
       97.6$\scriptstyle\pm 0.1$  &
       66.4$\scriptstyle\pm 0.4$  &
       74.0$\scriptstyle\pm 0.6$  &
       77.8$\scriptstyle\pm 0.7$  &
       79.0  \\
        \hline
        MixStyle \cite{zhou2021domain} &    
        \multirow{7}{*}[0.5em]{\emph{Data Augmentation}} &   
        98.6$\scriptstyle\pm 0.3$  &
        64.5$\scriptstyle\pm 1.1$  &
        72.6$\scriptstyle\pm 0.5$  &
        75.7$\scriptstyle\pm 1.7$  &
        77.9  \\
        RSC \cite{huang2020self} &
        {} &
        97.9$\scriptstyle\pm 0.1$  &
        62.5$\scriptstyle\pm 0.7$  &
        72.3$\scriptstyle\pm 1.2$  &
        75.6$\scriptstyle\pm 0.8$  &
        77.1 \\
        Mixup \cite{yan2020improve} &
        {} &
        98.3$\scriptstyle\pm 0.6$  &
        64.8$\scriptstyle\pm 1.0$  &
        72.1$\scriptstyle\pm 0.5$  &
        74.3$\scriptstyle\pm 0.8$  &
        77.4  \\
        SagNets \cite{nam2021reducing} &
        {} &
        97.9$\scriptstyle\pm 0.4$  &
        64.5$\scriptstyle\pm 0.5$  &
        71.4$\scriptstyle\pm 1.3$  &
        77.5$\scriptstyle\pm 0.5$  &
        77.8 \\
        \rowcolor[HTML]{EFEFEF}
        DCAug$^{domain}$ (Ours) &
        {} &
        98.3$\scriptstyle\pm 0.3$  &
        64.7$\scriptstyle\pm 0.2$  &
        74.2$\scriptstyle\pm 0.6$  &
        78.3$\scriptstyle\pm 0.8$  &
        78.9 \\
        \rowcolor[HTML]{EFEFEF}
        DCAug$^{label}$ (Ours) &
        {} &
        98.3$\scriptstyle\pm 0.1$ &
        64.2$\scriptstyle\pm 0.4$ &
        74.4$\scriptstyle\pm 0.6$ &
        77.5$\scriptstyle\pm 0.3$  &
        78.6  \\
        \rowcolor[HTML]{EFEFEF}
        TeachDCAug$^{label}$ (Ours) &
        {} &
        98.5$\scriptstyle\pm 0.1$ &
        63.7$\scriptstyle\pm 0.3$ &
        75.6$\scriptstyle\pm 0.5$ &
        77.0$\scriptstyle\pm 0.7$  &
        78.7  \\
        \bottomrule
    \end{tabular}
   \caption{Out-of-domain accuracies (\%) on VLCS.}
   \label{tab:vlcs}
\end{table*}

\begin{table*}[h]
    \centering
    \begin{tabular}{@{}lcccccc@{}}

        \toprule
        \multirow{2}{*}[-1em]{\textbf{Method}} & \multirow{2}{*}[-1em]{\textbf{Category}} & {} & {} & {} &
        \\
        {} & {} & {} & \multicolumn{2}{c}{\multirow{2}{*}[1em]{$\qquad$$\qquad$\textbf{Domain}}} \\
        \cmidrule(lr){3-7}
        \addlinespace[5pt]
        {} & {} & Art & Clipart & Product & Real & \textbf{Avg.}\\
        \midrule
        ERM \cite{vapnik1991principles} &
        \emph{Baseline} &
        63.1$\scriptstyle\pm 0.3$  &
        51.9$\scriptstyle\pm 0.4$  &
        77.2$\scriptstyle\pm 0.5$  &
        78.1$\scriptstyle\pm 0.2$  &
        67.6 \\
        \hline
        
        MMD \cite{li2018domain} &
        \multirow{4}{*}[-1em]{\emph{Domain-Invariant}} &
        60.4$\scriptstyle\pm 0.2$  &
        53.3$\scriptstyle\pm 0.3$  &
        74.3$\scriptstyle\pm 0.1$  &
        77.4$\scriptstyle\pm 0.6$  &
        66.4  \\
        IRM \cite{arjovsky2019invariant} &
        {} &
        58.9$\scriptstyle\pm 2.3$  &
        52.2$\scriptstyle\pm 1.6$  &
        72.1$\scriptstyle\pm 2.9$  &
        74.0$\scriptstyle\pm 2.5$  &
        64.3  \\
        GroupDRO \cite{sagawa2019distributionally} &
        {} &
        60.4$\scriptstyle\pm 0.7$  &
        52.7$\scriptstyle\pm 1.0$  &
        75.0$\scriptstyle\pm 0.7$  &
        76.0$\scriptstyle\pm 0.7$  &
        66.0  \\
        DANN \cite{ganin2016domain} &
        {} &
        59.9$\scriptstyle\pm 1.3$  &
        53.0$\scriptstyle\pm 0.3$  &
        73.6$\scriptstyle\pm 0.7$  &
        76.9$\scriptstyle\pm 0.5$  &
        65.9  \\
        CORAL \cite{sun2016deep} &
        {} &
        65.3$\scriptstyle\pm 0.4$  &
        54.4$\scriptstyle\pm 0.5$  &
        76.5$\scriptstyle\pm 0.1$  &
        78.4$\scriptstyle\pm 0.5$  &
        68.7  \\
        mDSDI \cite{bui2021exploiting} &
        {} &
        68.1$\scriptstyle\pm 0.3$  &
        52.1$\scriptstyle\pm 0.4$  &
        76.0$\scriptstyle\pm 0.2$  &
        80.4$\scriptstyle\pm 0.2$  &
        69.2  \\
        
        \hline
        DDAIG \cite{zhou2020deep} &
        \multirow{8}{*}[0.5em]{\emph{Data Augmentation}} & 
        59.2  &
        52.3  &
        74.6  &
        76.0  &
        65.5  \\
        MixStyle \cite{zhou2021domain} &
        {} &
        51.1$\scriptstyle\pm 0.3$  &
        53.2$\scriptstyle\pm 0.4$  &
        68.2$\scriptstyle\pm 0.7$  &
        69.2$\scriptstyle\pm 0.6$  &
        60.4  \\
        RSC \cite{huang2020self} &
        {} &
        60.7$\scriptstyle\pm 1.4$  &
        51.4$\scriptstyle\pm 0.3$  &
        74.8$\scriptstyle\pm 1.1$  &
        75.1$\scriptstyle\pm 1.3$  &
        65.5 \\
        Mixup \cite{yan2020improve} &
        {} &
        62.4$\scriptstyle\pm 0.8$  &
        54.8$\scriptstyle\pm 0.6$  &
        76.9$\scriptstyle\pm 0.3$  &
        78.3$\scriptstyle\pm 0.2$  &
        68.1  \\
        SagNets \cite{nam2021reducing} &
        {} &
        63.4$\scriptstyle\pm 0.2$  &
        54.8$\scriptstyle\pm 0.4$  &
        75.8$\scriptstyle\pm 0.4$  &
        78.3$\scriptstyle\pm 0.3$  &
        68.1 \\
        \rowcolor[HTML]{EFEFEF}
        DCAug$^{domain}$ (Ours) &
        {} &
        62.4$\scriptstyle\pm 0.4$  &
        56.7$\scriptstyle\pm 0.5$  &
        77.0$\scriptstyle\pm 0.4$  &
        79.0$\scriptstyle\pm 0.1$  &
        68.8 \\
        \rowcolor[HTML]{EFEFEF}
        DCAug$^{label}$ (Ours) &
        {} &
        61.8$\scriptstyle\pm 0.6$ &
        55.4$\scriptstyle\pm 0.6$ &
        77.1$\scriptstyle\pm 0.3$ &
        78.9$\scriptstyle\pm 0.3$  &
        68.3  \\
        \rowcolor[HTML]{EFEFEF}
        TeachDCAug$^{label}$ (Ours) &
        {} &
        66.2$\scriptstyle\pm 0.2$ &
        57.0$\scriptstyle\pm 0.3$ &
        78.3$\scriptstyle\pm 0.1$ &
        80.1$\scriptstyle\pm 0.0$  &
        70.4  \\
        
        \bottomrule
    \end{tabular}
   \caption{Out-of-domain accuracies (\%) on OfficeHome.}
   \label{tab:office}
\end{table*}

\begin{table*}[h]
    \centering
    \begin{tabular}{@{}lcccccc@{}}

        \toprule
        \multirow{2}{*}[-1em]{\textbf{Method}} & \multirow{2}{*}[-1em]{\textbf{Category}} & {} & {} & {} &
        \\
        {} & {} & {} & \multicolumn{2}{c}{\multirow{2}{*}[1em]{$\qquad$$\qquad$\textbf{Domain}}} \\
        \cmidrule(lr){3-7}
        \addlinespace[5pt]
        {} & {} & L100 & L38 & L43 & L46 & \textbf{Avg.}\\
        \midrule
        ERM \cite{vapnik1991principles} &
        \emph{Baseline} &
        54.3$\scriptstyle\pm 0.4$  &
        42.5$\scriptstyle\pm 0.7$  &
        55.6$\scriptstyle\pm 0.3$  &
        38.8$\scriptstyle\pm 2.5$  &
        47.8 \\
        \hline
        
        MMD \cite{li2018domain} &
        \multirow{4}{*}[-1em]{\emph{Domain-Invariant}} &
        41.9$\scriptstyle\pm 3.0$  &
        34.8$\scriptstyle\pm 1.0$  &
        57.0$\scriptstyle\pm 1.9$  &
        35.2$\scriptstyle\pm 1.8$  &
        42.2 \\
        IRM \cite{arjovsky2019invariant} &
        {} &
        54.6$\scriptstyle\pm 1.3$  &
        39.8$\scriptstyle\pm 1.9$  &
        56.2$\scriptstyle\pm 1.8$  &
        39.6$\scriptstyle\pm 0.8$  &
        47.6 \\
        GroupDRO \cite{sagawa2019distributionally} &
        {} &
        41.2$\scriptstyle\pm 0.7$  &
        38.6$\scriptstyle\pm 2.1$  &
        56.7$\scriptstyle\pm 0.9$  &
        36.4$\scriptstyle\pm 2.1$  &
        43.2 \\
        DANN \cite{ganin2016domain} &
        {} &
        51.1$\scriptstyle\pm 3.5$  &
        40.6$\scriptstyle\pm 0.6$  &
        57.4$\scriptstyle\pm 0.5$  &
        37.7$\scriptstyle\pm 1.8$  &
        46.7 \\
        CORAL \cite{sun2016deep} &
        {} &
        51.6$\scriptstyle\pm 2.4$  &
        42.2$\scriptstyle\pm 1.0$  &
        57.0$\scriptstyle\pm 1.0$  &
        39.8$\scriptstyle\pm 2.9$  &
        47.7 \\
        mDSDI \cite{bui2021exploiting} &
        {} &
        53.2$\scriptstyle\pm 3.0$  &
        43.3$\scriptstyle\pm 1.0$  &
        56.7$\scriptstyle\pm 0.5$  &
        39.2$\scriptstyle\pm 1.3$  &
        48.1 \\
        
        \hline
        MixStyle \cite{zhou2021domain} &
        {} &
        54.3$\scriptstyle\pm 1.1$  &
        34.1$\scriptstyle\pm 1.1$  &
        55.9$\scriptstyle\pm 1.1$  &
        31.7$\scriptstyle\pm 2.1$  &
        44.0 \\
        RSC \cite{huang2020self} &
        {} &
        50.2$\scriptstyle\pm 2.2$  &
        39.2$\scriptstyle\pm 1.4$  &
        56.3$\scriptstyle\pm 1.4$  &
        40.8$\scriptstyle\pm 0.6$  &
        46.6 \\
        Mixup \cite{yan2020improve} &
        {} &
        59.6$\scriptstyle\pm 2.0$  &
        42.2$\scriptstyle\pm 1.4$  &
        55.9$\scriptstyle\pm 0.8$  &
        33.9$\scriptstyle\pm 1.4$  &
        47.9 \\
        SagNets \cite{nam2021reducing} &
        {} &
        53.0$\scriptstyle\pm 2.9$  &
        43.0$\scriptstyle\pm 2.5$  &
        57.9$\scriptstyle\pm 0.6$  &
        40.4$\scriptstyle\pm 1.3$  &
        48.6 \\
        \rowcolor[HTML]{EFEFEF}
        DCAug$^{domain}$ (Ours) &
        {} &
        59.0$\scriptstyle\pm 0.5$  &
        42.7$\scriptstyle\pm 1.1$  &
        54.2$\scriptstyle\pm 1.5$  &
        38.9$\scriptstyle\pm 0.2$  &
        48.7 \\
        \rowcolor[HTML]{EFEFEF}
        DCAug$^{label}$ (Ours) &
        {} &
        56.1$\scriptstyle\pm 1.3$ &
        44.5$\scriptstyle\pm 1.7$ &
        57.1$\scriptstyle\pm 1.3$ &
        39.4$\scriptstyle\pm 1.7$  &
        49.3  \\
        \rowcolor[HTML]{EFEFEF}
        TeachDCAug$^{label}$ (Ours) &
        {} &
        60.6$\scriptstyle\pm 0.6$ &
        43.0$\scriptstyle\pm 2.0$ &
        58.5$\scriptstyle\pm 0.3$ &
        42.3$\scriptstyle\pm 1.4$  &
        51.1  \\
        
        \bottomrule
    \end{tabular}
   \caption{Out-of-domain accuracies (\%) on TerraIncognita.}
   \label{tab:terra}
\end{table*}

\begin{table*}[h]
    \centering
    \resizebox{\linewidth}{!}{%
    \begin{tabular}{@{}lcccccccc@{}}
    
        \toprule
        \multirow{2}{*}[-1em]{\textbf{Method}} & \multirow{2}{*}[-1em]{\textbf{Category}} & {} & {} & {} &
        \\
        {} & {} & {} & \multicolumn{4}{c}{\multirow{2}{*}[1em]{$\qquad$$\qquad$\textbf{Domain}}} \\
        \cmidrule(lr){3-9}
        \addlinespace[5pt]

        {} & {} & Clipart & Infograph & Painting & Quickdraw & Real &  Sketch & \textbf{Avg.}\\
        \midrule
        ERM \cite{vapnik1991principles} &
        \emph{Baseline} &
        63.0$\scriptstyle\pm 0.2$  &
        21.2$\scriptstyle\pm 0.2$  &
        50.1$\scriptstyle\pm 0.4$  &
        13.9$\scriptstyle\pm 0.5$  &
        63.7$\scriptstyle\pm 0.2$  &
        52.0$\scriptstyle\pm 0.5$  &
        44.0 \\
        \hline
        
        MMD \cite{li2018domain} &
        \multirow{4}{*}[-1em]{\emph{Domain-Invariant}} &
        32.1$\scriptstyle\pm 13.3$  &
        11.0$\scriptstyle\pm 4.6$  &
        26.8$\scriptstyle\pm 11.3$  &
        8.7 $\scriptstyle\pm 2.1$  &
        32.7$\scriptstyle\pm 13.8$  &
        28.9$\scriptstyle\pm 11.9$  &
        23.4  \\
        IRM \cite{arjovsky2019invariant} &
        & {}
        48.5$\scriptstyle\pm 2.8$  &
        15.0$\scriptstyle\pm 1.5$  &
        38.3$\scriptstyle\pm 4.3$  &
        10.9$\scriptstyle\pm 0.5$  &
        48.2$\scriptstyle\pm 5.2$  &
        42.3$\scriptstyle\pm 1.1$  &
        33.9  \\
        GroupDRO \cite{sagawa2019distributionally} &
        & {}
        42.7$\scriptstyle\pm 0.5$  &
        17.5$\scriptstyle\pm 0.4$  &
        33.8$\scriptstyle\pm 0.5$  &
        9.3$\scriptstyle\pm 0.3$  &
        51.6$\scriptstyle\pm 0.4$  &
        40.1$\scriptstyle\pm 0.6$  &
        33.3  \\
        DANN \cite{ganin2016domain} &
        & {}
        53.1$\scriptstyle\pm 0.2$  &
        18.3$\scriptstyle\pm 0.1$  &
        44.2$\scriptstyle\pm 0.7$  &
        11.8$\scriptstyle\pm 0.1$  &
        55.5$\scriptstyle\pm 0.4$  &
        46.8$\scriptstyle\pm 0.6$  &
        38.3  \\
        CORAL \cite{sun2016deep} &
        & {}
        59.2$\scriptstyle\pm 0.1$  &
        19.7$\scriptstyle\pm 0.2$  &
        46.6$\scriptstyle\pm 0.3$  &
        13.4$\scriptstyle\pm 0.4$  &
        59.8$\scriptstyle\pm 0.2$  &
        50.1$\scriptstyle\pm 0.6$  &
        41.5  \\
        mDSDI \cite{bui2021exploiting} &
        & {}
        62.1$\scriptstyle\pm 0.3$  &
        19.1$\scriptstyle\pm 0.4$  &
        49.4$\scriptstyle\pm 0.4$  &
        12.8$\scriptstyle\pm 0.7$  &
        62.9$\scriptstyle\pm 0.3$  &
        50.4$\scriptstyle\pm 0.4$  &
        42.8  \\
        \hline        
        MixStyle \cite{zhou2021domain} &
        \multirow{7}{*}[0.5em]{\emph{Data Augmentation}} & 
        51.9$\scriptstyle\pm 0.4$  &
        13.3$\scriptstyle\pm 0.2$  &
        37.0$\scriptstyle\pm 0.5$  &
        12.3$\scriptstyle\pm 0.1$  &
        46.1$\scriptstyle\pm 0.3$  &
        43.4$\scriptstyle\pm 0.4$  &
        34.0  \\
        RSC \cite{huang2020self} &
        & {}
        55.0$\scriptstyle\pm 1.2$  &
        18.3$\scriptstyle\pm 0.5$  &
        44.4$\scriptstyle\pm 0.6$  &
        12.2$\scriptstyle\pm 0.2$  &
        55.7$\scriptstyle\pm 0.7$  &
        47.8$\scriptstyle\pm 0.9$  &
        38.9 \\
        Mixup \cite{yan2020improve} &
        & {}
        55.7$\scriptstyle\pm 0.3$  &
        18.5$\scriptstyle\pm 0.5$  &
        44.3$\scriptstyle\pm 0.5$  &
        12.5$\scriptstyle\pm 0.4$  &
        55.8$\scriptstyle\pm 0.3$  &
        48.2$\scriptstyle\pm 0.5$  &
        39.2  \\
        SagNets \cite{nam2021reducing} &
        & {}
        57.7$\scriptstyle\pm 0.3$  &
        19.0$\scriptstyle\pm 0.2$  &
        45.3$\scriptstyle\pm 0.3$  &
        12.7$\scriptstyle\pm 0.5$  &
        58.1$\scriptstyle\pm 0.5$  &
        48.8$\scriptstyle\pm 0.2$  &
        40.3 \\
        \rowcolor[HTML]{EFEFEF}
        DCAug$^{domain}$ (Ours) &
        & {}
        62.8$\scriptstyle\pm 0.2$  &
        19.9$\scriptstyle\pm 0.2$  &
        50.6$\scriptstyle\pm 0.3$  &
        13.5$\scriptstyle\pm 0.3$  &
        63.0$\scriptstyle\pm 0.1$  &
        52.3$\scriptstyle\pm 0.4$  &
        43.7 \\
        \rowcolor[HTML]{EFEFEF}
        DCAug$^{label}$ (Ours) &
        & {}
        62.5$\scriptstyle\pm 0.2$ &
        20.0$\scriptstyle\pm 0.2$ &
        50.4$\scriptstyle\pm 0.1$ &
        13.9$\scriptstyle\pm 0.3$ &
        62.9$\scriptstyle\pm 0.2$ &
        53.2$\scriptstyle\pm 0.4$ &
        43.8  \\
        \rowcolor[HTML]{EFEFEF}
        TeachDCAug$^{label}$ (Ours) &
        & {}
        65.5$\scriptstyle\pm 0.0$ &
        22.2$\scriptstyle\pm 0.0$ &
        53.7$\scriptstyle\pm 0.0$ &
        15.6$\scriptstyle\pm 0.1$ &
        65.8$\scriptstyle\pm 0.1$ &
        55.9$\scriptstyle\pm 0.1$ &
        46.4  \\
        
        \bottomrule
    \end{tabular}}
   \caption{Out-of-domain accuracies (\%) on DomainNet.}
   \label{tab:domainnet}
\end{table*}

\end{appendices}

\end{document}